\documentclass[runningheads]{llncs}

\usepackage{amsmath}
\usepackage{amssymb}
\usepackage{booktabs}
\usepackage{multirow}
\usepackage{multicol}
\usepackage{pifont}
\usepackage{bm}
\usepackage{algpseudocode}
\usepackage{wrapfig}
\usepackage{subfigure}
\usepackage{graphicx}

\usepackage{bbm}
\usepackage{makecell}
\usepackage{caption}
\usepackage{xcolor}
\usepackage[linesnumbered,ruled]{algorithm2e}
\usepackage{hyperref}
\usepackage{marvosym}
    
\usepackage{tikz}
\usepackage{comment}
\usepackage{amsmath,amssymb} 
\usepackage{color}
\usepackage{wrapfig}
\usepackage{xcolor}

\usepackage[accsupp]{axessibility}  

\begin{document}
\pagestyle{headings}
\mainmatter
\def\ECCVSubNumber{5837}  

\title{Hardly Perceptible Trojan Attack against Neural Networks with Bit Flips} 

\titlerunning{Hardly Perceptible Trojan Attack against Neural Networks with Bit Flips}
%
\author{Jiawang Bai\inst{1} \and
Kuofeng Gao\inst{1} \and
Dihong Gong\inst{2} \and 
Shu-Tao Xia\inst{1,3, {{\textrm{\Letter}}}} \and \\
Zhifeng Li\inst{2, {{\textrm{\Letter}}}} \and
Wei Liu\inst{2, {{\textrm{\Letter}}}}}
%

\authorrunning{Jiawang Bai, Kuofeng Gao, Dihong Gong et al.}

%
\institute{Tsinghua Shenzhen International Graduate School, Tsinghua University \and
Data Platform, Tencent \and 
Research Center of Artificial Intelligence, Peng Cheng Laboratory\
\email{\{bjw19,gkf21\}@mails.tsinghua.edu.cn;} \email{gongdihong@gmail.com;} \email{xiast@sz.tsinghua.edu.cn;} \email{michaelzfli@tencent.com;} \email{wl2223@columbia.edu}}
\renewcommand{\thefootnote}{\fnsymbol{footnote}}
\footnotetext{\textrm{\Letter} Corresponding authors.}
\maketitle

\begin{abstract}
The security of deep neural networks (DNNs) has attracted increasing attention due to their widespread use in various applications. Recently, the deployed DNNs have been demonstrated to be vulnerable to Trojan attacks, which manipulate model parameters with bit flips to inject a hidden behavior and activate it by a specific trigger pattern. However, all existing Trojan attacks adopt noticeable patch-based triggers (e.g., a square pattern), making them perceptible to humans and easy to be spotted by machines. In this paper, we present a novel attack, namely \underline{h}ardly \underline{p}erceptible \underline{T}rojan attack (HPT). HPT crafts hardly perceptible Trojan images by utilizing the additive noise and per-pixel flow field to tweak the pixel values and positions of the original images, respectively. To achieve superior attack performance, we propose to jointly optimize bit flips, additive noise, and flow field. Since the weight bits of the DNNs are binary, this problem is very hard to be solved. We handle the binary constraint with equivalent replacement and provide an effective optimization algorithm. Extensive experiments on CIFAR-10, SVHN, and ImageNet datasets show that the proposed HPT can generate hardly perceptible Trojan images, while achieving comparable or better attack performance compared to the state-of-the-art methods.
The code is available at: \href{https://github.com/jiawangbai/HPT}{\texttt{https://github.com/jiawangbai/HPT}}.

\end{abstract}

\section{Introduction}
\label{sec:intro}

Although deep neural networks (DNNs) have been showing state-of-the-art performances in various complex tasks, such as image classification \cite{simonyan2014very,russakovsky2015imagenet,he2016deep}, facial recognition \cite{tang2004video,li2014common,deng2019mutual,qiu2021end2end}, and object detection \cite{girshick2015fast,redmon2016you}, prior studies have revealed their vulnerability against diverse attacks \cite{goodfellow2014explaining,moosavi2017universal,shafahi2018poison,gu2019badnets,nguyen2020input,li2022backdoor,saha2020hidden,souri2021sleeper,chen2022adversarial}. One such attack is the \textit{Trojan attack} \cite{LiuMALZW018} happening in the deployment stage, in which an attacker manipulates a DNN to inject a hidden behavior called Trojan. The Trojan can only be activated by the specific trigger pattern.

Trojan attacks on a deployed DNN alter the model parameters in the memory using bit flip techniques, e.g., Row Hammer Attack \cite{kim2014flipping,van2016drammer}, but do not tamper with the training pipeline 
and have no extra forward or backward calculation during inference. Then, the attacked DNN makes a target prediction on the inputs with the trigger, while behaving normally on clean samples \cite{LiuMALZW018,rakin2020tbt,chen2021proflip}. These dangerous properties pose severe threats to DNN-based applications after model deployment. Therefore, it is necessary to study the Trojan attacks on the deployed DNNs in order to recognize their flaws and solve related security risks.



\begin{figure}[t]
    \begin{minipage}[b]{0.57\textwidth}
    \centering
    \includegraphics[width=\textwidth]{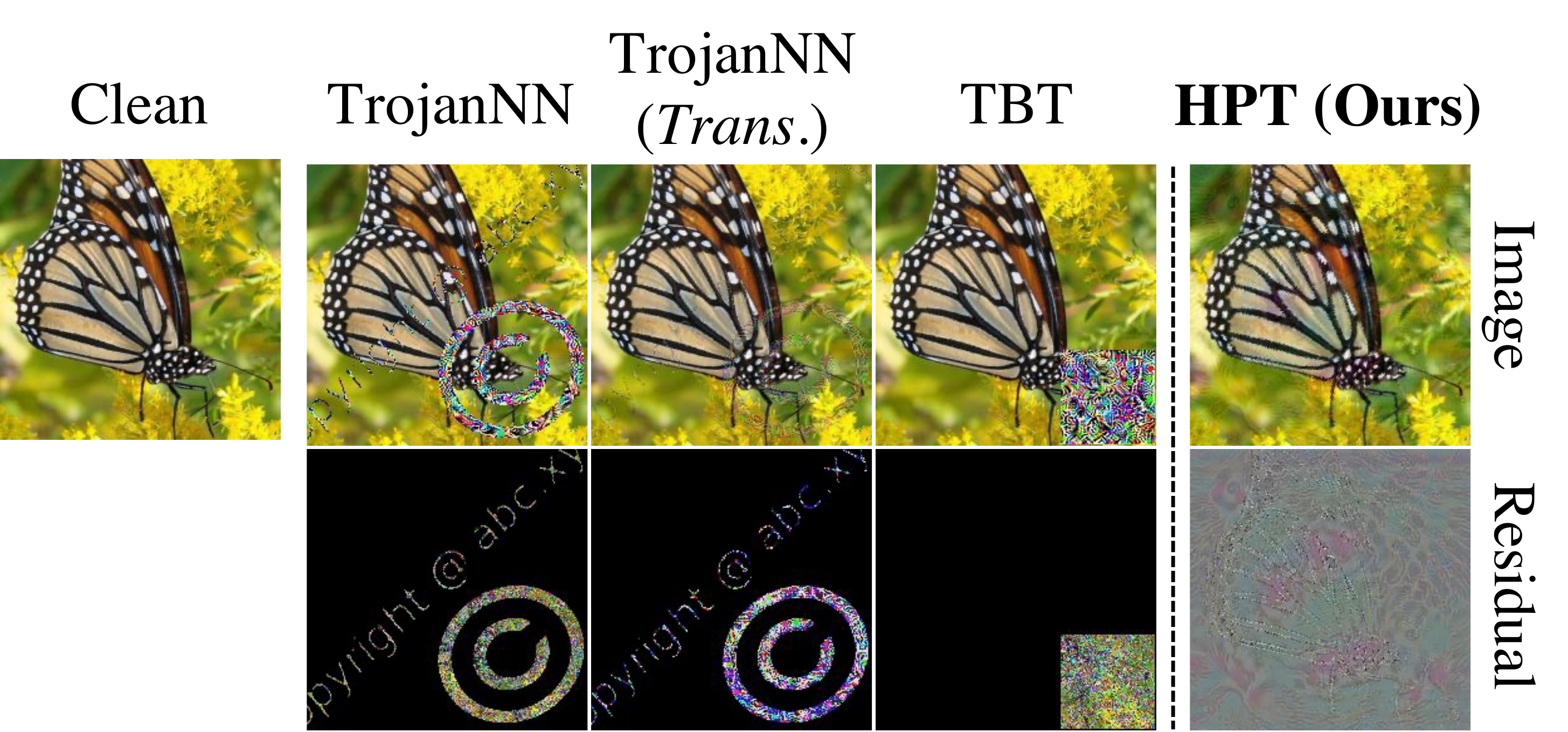}
    \end{minipage}%
    \hfill
    \begin{minipage}[b]{0.4\textwidth}
    \centering
    
\setlength{\abovecaptionskip}{-1pt}
\setlength{\belowcaptionskip}{-30pt}
    \caption{{Visualization of Trojan images from TrojanNN \cite{LiuMALZW018}, TBT \cite{rakin2020tbt}, and HPT. \textit{`Trans.'} indicates the use of transparent trigger proposed in \cite{LiuMALZW018}. Note that ProFlip \cite{chen2021proflip} uses the same trigger pattern (a square pattern) as TBT.}}
	\label{fig:vis_intro}
    \end{minipage}
\end{figure}

A Trojan attack generally is composed of two subroutines: critical bits identification and specific trigger generation. Previous works \cite{rakin2019bit,chen2021proflip} made efforts towards reducing the number of bit flips by developing search strategies. After flipping the identified bits, the attacker can activate the hidden behavior using the Trojan images, which are any images embedded with a specific patch-based trigger, such as a watermark pattern in \cite{LiuMALZW018} or a square pattern in \cite{rakin2020tbt,chen2021proflip}. However, due to these unnatural and noticeable triggers, the Trojan images can be easily spotted by humans \cite{nguyen2020wanet,doan2021lira} and machines. 
For example, we construct a simple linear classifier to distinguish clean images and Trojan images crafted by TBT \cite{rakin2020tbt} based on their Grad-CAM heatmaps \cite{selvaraju2017grad} on ImageNet, resulting in a 98.0\% success rate.
The transparent trigger \cite{LiuMALZW018} was proposed to reduce the perceptibility, but leading to lower attack performance. Hence, how to inject less perceptible Trojan with superior attack performance is a challenging problem.

To address the aforementioned problems, we propose the hardly perceptible Trojan attack (HPT).
Instead of applying the patch-based trigger predefined by a mask, HPT tweaks the pixel values and positions of the original images to craft Trojan images. Specifically, we modify pixel values by adding the pixel-wise noise inspired by adversarial examples \cite{szegedy2013intriguing,goodfellow2014explaining,moosavi2017universal} and change pixel positions by per-pixel flow field \cite{duchon1977splines,jaderberg2015spatial,zhou2016view}. 
As shown in Figure \ref{fig:vis_intro}, Trojan images of HPT are less perceptible and harder to be distinguished from original images. It will be further demonstrated in the human perceptual study in Section \ref{sec:human}.


Since the value of each weight bit is `0' or `1', we cast each bit as a binary variable. Integrating Trojan images generation and critical bits identification, we formulate the proposed HPT as a mixed integer programming (MIP) problem, i.e.,  jointly optimizing bit flips, additive noise, and flow field. Moreover, we constrain the modification on image pixels and the number of bit flips to yield a hardly perceptible and practical attack \cite{van2016drammer,zhao2019fault}.
To solve this MIP problem, we reformulate it as an equivalent continuous problem \cite{wu2018ell} and present an effective optimization algorithm based on the standard alternating direction method of multipliers (ADMM) \cite{gabay1976dual,boyd2011distributed}. We conduct extensive experiments on CIFAR-10, SVHN, and ImageNet with 8-bit quantized ResNet-18 and VGG-16 architectures following \cite{rakin2020tbt,chen2021proflip}, which shows that HPT is hardly perceptible and achieves superior attack performance. The attributes of compared methods and HPT are summarized in Table \ref{tab:attributes}.

The contributions are summarized as follows:
\begin{itemize}
	\item
	For the first time, we improve Trojan attacks on the deployed DNNs to be strong and hardly perceptible. We investigate the use of modifying the pixel values and positions of the original images to craft Trojan images.  
	\item We formulate the proposed HPT as a constrained MIP problem to jointly optimize bit flips, additive noise, and flow field, and further provide an effective optimization algorithm.
	\item Finally, HPT obtains both hardly perceptible Trojan images and promising attack performance, e.g., an attack success rate of 95\% with only 10 bit flips out of 88 million bits in attacking ResNet-18 on ImageNet.
\end{itemize}

\begin{table}[t]
\centering
\caption{Summary of attributes of TrojanNN \cite{LiuMALZW018}, TBT \cite{rakin2020tbt}, ProFlip \cite{chen2021proflip}, and HPT. \textit{`Trans.'} indicates the use of transparent trigger proposed in \cite{LiuMALZW018} and `ASR' denotes the attack success rate.}
\setlength\tabcolsep{7pt}
\resizebox{0.9\linewidth}{!}{
\begin{tabular}{lccc}
\hline
\textbf{Method} & \textbf{High ASR} & \textbf{\begin{tabular}[c]{@{}c@{}}A Small Set  of Bit  Flips\end{tabular}} & \textbf{\begin{tabular}[c]{@{}c@{}}Hardly Perceptible Trigger\end{tabular}} \\ \hline
TrojanNN & \checkmark &  &  \\
\multicolumn{1}{l}{TrojanNN ($Trans.$)} & \multicolumn{1}{l}{} & \multicolumn{1}{l}{} & \checkmark \\
TBT  & \checkmark &  &  \\
ProFlip & \checkmark & \checkmark &  \\
\textbf{HPT (Ours)} & \checkmark & \checkmark & \checkmark \\ \hline
\end{tabular}
}
\label{tab:attributes}
\end{table}

\section{Related Works and Preliminaries}

\noindent \textbf{Attacks on the Deployed DNNs.}
Recently, since DNNs have been widely applied to security-critical tasks, e.g., facial
recognition \cite{gong2013multi,wen2016discriminative,wang2018cosface,yang2021larnet}, their security in the deployment stage has received extensive attention. Previous works assume that the attacker can modify the weight parameters of the deployed DNNs to achieve some malicious purposes, e.g., misclassifying certain samples \cite{liu2017fault,zhao2019fault}. 
Some physical techniques (e.g., Row Hammer Attack \cite{kim2014flipping,van2016drammer} and Laser Beam Attack \cite{agoyan2010flip,colombier2019laser}) can precisely flip bits (`0'$\rightarrow$`1' or `1'$\rightarrow$`0') in the memory without accessing them. These techniques allow an attacker to attack a deployed DNN by modifying its bit representation \cite{YaoRF20,venceslai2020neuroattack,rakin2021t}. For instance, Rakin et al. \cite{rakin2019bit} presented that an extremely small number of bit flips can crush a fully functional DNN to a random output generator. After that, Bai et al. \cite{bai2021targeted} proposed to attack a specified sample into a target class via flipping limited bits. To mitigate the bit flip-based attacks, some defense mechanisms \cite{he2020defending,li2020defending,liu2020concurrent,li2021radar} have been explored.

As a line of research, \textit{Trojan attacks} \cite{rakin2020tbt,chen2021proflip} insert a hidden behavior in the DNN using bit flip techniques, which can be activated by a designed trigger. Specifically, an attacked DNN will make wrong predictions on the inputs with the presence of the trigger, while behaving normally on original samples. Previous works proposed to generate a specified pattern as the trigger (e.g., a square pattern) \cite{LiuMALZW018,rakin2020tbt,chen2021proflip}. Besides, to reduce the number of bit flips, heuristic strategies are designed to identify the critical bits, e.g., neural gradient ranking in \cite{rakin2020tbt} or progressive search algorithm in \cite{chen2021proflip}. The results in \cite{chen2021proflip} show that only a few bit flips yield a high attack success rate, which further raises the security concerns. 

\noindent \textbf{Quantized DNNs.} 
In the deployment stage, model quantization has been widely adopted to reduce the storage requirements and accelerate the inference speed \cite{lin2016fixed,krishnamoorthi2018quantizing}. In this paper, we adopt a layer-wise $Q$-bit uniform quantizer, which is identical to the Tensor-RT solution \cite{migacz20178}. Given binary weight parameters $\bm{\theta} \in \{0,1\}^{N \times Q}$ of a $Q$-bit quantized DNN $g$, each parameter is represented and stored as a signed integer in two’s complement representation, i.e., $\bm{\theta}_i=[\bm{\theta}_{i,Q};...;\bm{\theta}_{i,1}] \in \{0,1\}^Q$. For the $l$-th layer with the step size $\Delta_l$, the binary representation $\bm{\theta}_i$ can be converted into a real number, as follows:
\begin{equation}
\bm{W}_i =(-2^{Q-1} \cdot \bm{\theta}_{i,Q} + \sum_{j=1}^{Q-1}{2^{j-1} \cdot \bm{\theta}_{i,j} }) \cdot \Delta_l,
\label{eq:quantization}
\end{equation}
where $\bm{W} \in \mathbb{R}^{N}$ denotes the floating-point weight parameters of the DNN. For clarity, hereafter $\bm{\theta}$ is reshaped from the tensor to the vector, i.e., $\bm{\theta} \in \{0,1\}^{NQ}$.

\noindent \textbf{Threat Model.}
We consider the threat model used by previous bit flip-based Trojan attack studies \cite{rakin2020tbt,chen2021proflip}. The attacker knows the location of the victim DNN in the memory to implement precisely bit flips. We also assume that the attacker has a small set of clean data and knows the architecture and parameters of the victim DNN. Note that our attack does not have access to the training process nor the training data. During inference, the attacker can activate the injected Trojan by applying the generated trigger on the test samples.

\begin{figure*}[t]
  \centering
  \includegraphics[width=\linewidth]{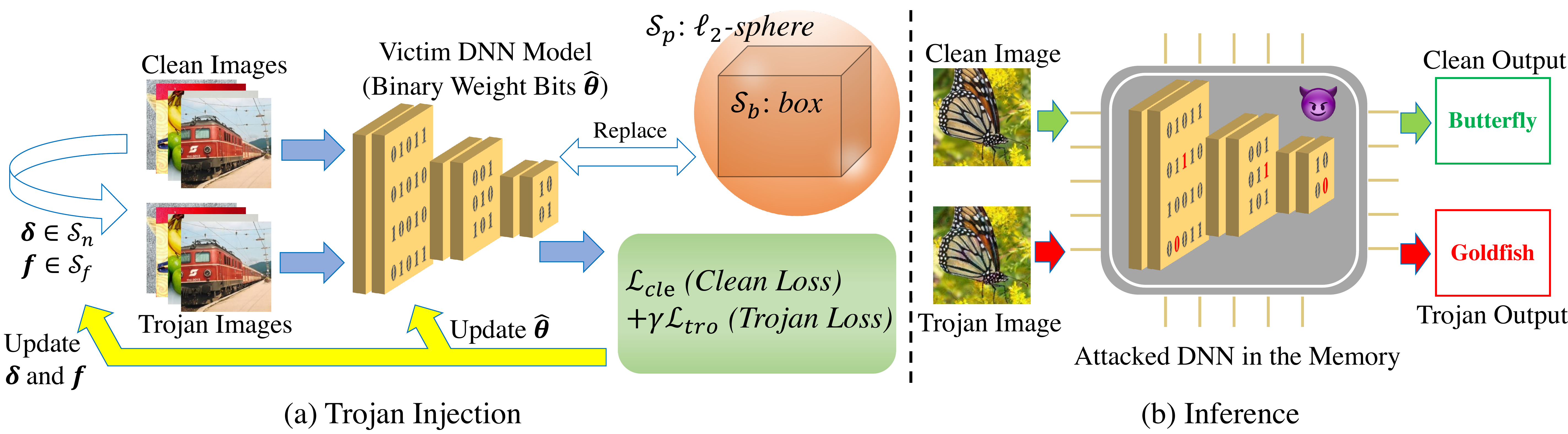}

\setlength{\abovecaptionskip}{-0pt}
\setlength{\belowcaptionskip}{-10pt}
  \caption{Pipeline of the proposed HPT, where the target class is `\textit{Goldfish}'. (a) Trojan Injection: optimizing $\hat{\bm{\theta}}$, $\bm{\delta}$, and $\bm{f}$ jointly. (b) Inference: activating the hidden behaviour using the Trojan image.}
  \label{fig:pipeline}
\end{figure*}

\section{Methodology}
In this section, we firstly describe the hardly perceptible trigger used by HPT. We then introduce the problem formulation and present an effective optimization algorithm based on ADMM. 
Figure \ref{fig:pipeline} shows the entire pipeline of HPT.

\subsection{Hardly Perceptible Trigger}
Let $\bm{x}\in \mathcal{X}$ denote a clean image, and $\hat{\bm{x}}$ denote its Trojan image, i.e.,  $\bm{x}$ with a specific trigger pattern. $\mathcal{X}=[0,1]^{HW \times C}$ is the image space, where $H$, $W$, $C$ are the height, width, and channel for an image, respectively. To modify the pixel values of the clean image, we apply additive noise $\bm{\delta}$ to the image $\bm{x}$. Inspired by the adversarial examples \cite{kurakin2016}, HPT requires $\bm{\delta} \in \mathcal{S}_n$, where $\mathcal{S}_n=\{\bm{\delta}:||\bm{\delta}||_\infty \leqslant \epsilon\}$ and $\epsilon$ denotes the maximum noise strength, so that $\hat{\bm{x}}$ is perceptually indistinguishable from $\bm{x}$.

To formulate changes of the pixel positions, we use $\bm{f} \in \mathbb{R}^{HW \times 2}$ to represent per-pixel flow field, where $\bm{f}^{(i)} = (\Delta u^{(i)}, \Delta v^{(i)})$ denotes the amount of displacement in each channel for each pixel $\hat{\bm{x}}^{(i)}$ within the Trojan image at the position $(\hat{u}^{(i)}, \hat{v}^{(i)})$.
Thus, the value of $\hat{\bm{x}}^{(i)}$ is sampled from position $(u^{(i)}, v^{(i)})=(\hat{u}^{(i)}+\Delta u^{(i)}, \hat{v}^{(i)}+\Delta v^{(i)})$ within the original image.  Since the sampled position is not necessarily an integer value, we use the differentiable bilinear interpolation \cite{zhou2016view} to generate the Trojan image considering four neighboring pixels around the location $(u^{(i)}, v^{(i)})$, denoted by $\mathcal{N}(u^{(i)}, v^{(i)})$.
To preserve high perceptual quality of the Trojan images, we enforce the local smoothness of the flow field $\bm{f}$ based on the total variation
\cite{xiao2018spatially,zhang2020generalizing}:
\begin{equation}
{
\mathcal{F}(\bm{f})=\!\sum_{p}^{all \ pixels}\!\!\! \sum_{q\in \mathcal{N}(p)}\!\!\! \sqrt{||\Delta u^{(p)}\!-\!\Delta u^{(q)}||_2^2\!+\!||\Delta v^{(p)}\!-\!\Delta v^{(q)}||_2^2}.
}
\label{eq:h}
\end{equation}
We introduce a hyper-parameter $\kappa$ to restrict $\mathcal{F}(\bm{f})$, i.e., $\bm{f} \in \mathcal{S}_f$ where $\mathcal{S}_f=\{\bm{f}: \mathcal{F}(\bm{f}) \leqslant \kappa\}$.

Based on the additive noise $\bm{\delta}$ and the flow field $\bm{f}$, each pixel before the clipping operation can be calculated as:
\begin{equation}
{
\hat{\bm{x}}^{(i)}=\!\sum_{\!\!q\in \mathcal{N}(u^{\!(i)}, v^{\!(i)})}\! (\bm{x}^{(q)}\!+\!\bm{\delta}^{(q)})(1\!-\!|u^{(i)}\!-\!u^{(q)}|)(1\!-\!|v^{(i)}\!-\!v^{(q)}|).
}
\label{eq:each_pixel}
\end{equation}
We can craft the Trojan image $\hat{\bm{x}}$ by calculating each pixel of $\hat{\bm{x}}$ according to Eqn. (\ref{eq:each_pixel}) and performing the $[0,1]$ clipping to ensure that it is in the image space, which is denoted as: 
\begin{equation}
\begin{split}
\hat{\bm{x}}=\mathcal{T}(\bm{x};\bm{\delta},\bm{f}).
\end{split}
\end{equation}
Note that $\hat{\bm{x}}$ is differentiable with respect to $\bm{\delta}$ and $\bm{f}$, enabling us to optimize them by the gradient method. We obtain $\bm{\delta}$ and $\bm{f}$ after the Trojan injection stage and apply them on any image to craft its Trojan image during inference.

\subsection{Problem Formulation}
Suppose that the victim DNN is the well-trained $Q$-bit quantized classifier $g:\mathcal{X} \rightarrow \mathcal{Y}$, where $\mathcal{Y}=\{1,...,K\}$ is the output space and $K$ is the number of classes. $\bm{\theta} \in \{0,1\}^{NQ}$ is the original binary weight parameters and $\hat{\bm{\theta}} \in \{0,1\}^{NQ}$ is the attacked parameters. As aforementioned, the attacker has a small set of clean data $D=\{(\bm{x}_i, y_i)\}_{i=1}^M$. To keep the attack stealthiness, we reduce the influence on original samples by minimizing the below loss:
{
\begin{equation}
\begin{split}
\mathcal{L}_{cle}(\hat{\bm{\theta}})=
\sum_{i=1}^{M} \ell(g(\bm{x}_i;\hat{\bm{\theta}}),y_i),
\end{split}
\end{equation}}
where $g_j(\bm{x}_i;\hat{\bm{\theta}})$ indicates the posterior probability of the input with respect to class $j$ and $\ell(\cdot,\cdot)$ denotes the cross-entropy loss.
Recall that the malicious purpose of Trojan attack is to classify the Trojan images into the target class $t$. 
To this end, we formulate this objective as:
{\setlength\abovedisplayskip{0.5em}
\setlength\belowdisplayskip{0.5em}
\begin{equation}
\begin{split}
\mathcal{L}_{tro}(\bm{\delta},\bm{f},\hat{\bm{\theta}})=
\sum_{i=1}^{M} \ell(g(\mathcal{T}(\bm{x}_i;\bm{\delta},\bm{f});\hat{\bm{\theta}}),t).
\end{split}
\label{eq:loss_tro}
\end{equation}}

Aligning with previous Trojan attacks \cite{rakin2020tbt,chen2021proflip}, reducing the number of bit flips is necessary to make the attack efficient and practical. HPT achieves this goal by restricting the Hamming distance between $\bm{\theta}$ and $\hat{\bm{\theta}}$ less than $b$. Considering the constraint on $\bm{\delta}$ and $\bm{f}$, we formulate the objective function of HPT as follows:
\begin{equation}
\begin{split}
\{\bm{\delta}^*,\bm{f}^*,& \hat{\bm{\theta}}^*\} = 
\mathrm{arg} \ \underset{\bm{\delta},\bm{f},\hat{\bm{\theta}}}{\mathrm{min}} \quad \mathcal{L}_{cle}(\hat{\bm{\theta}}) + \gamma \mathcal{L}_{tro}(\bm{\delta},\bm{f},\hat{\bm{\theta}}), \\
 s.t. \ \ \ & \bm{\delta} \in \mathcal{S}_n, \ \  \bm{f} \in \mathcal{S}_f, \ \
 \hat{\bm{\theta}} \in \{0,1\}^{N Q}, \ \ d_H(\bm{\theta},\hat{\bm{\theta}}) \leqslant b,
\end{split}
\label{eq:obj}
\end{equation}
where $\gamma$ is the hyper-parameter to balance the two terms and $d(\bm{\theta},\hat{\bm{\theta}})$ computes the Hamming distance between original and attacked parameters. Problem (\ref{eq:obj}) with the continuous variables $\bm{\delta}$ and $\bm{f}$ and the binary variable $\hat{\bm{\theta}}$ is a mixed integer programming, which is generally very difficult to solve. Here, inspired by a recent advanced work in integer programming \cite{wu2018ell}, we equivalently replace the binary constraint with the intersection of two continuous constraints, as presented in Proposition \ref{pro:lpbox}.
\begin{proposition}
\cite{wu2018ell} Let $\bm{1}_{NQ}$ denote the $NQ$-dimensional vector filled with all 1s. The binary set $\{0,1\}^{NQ}$ can be replaced by the intersection between $\mathcal{S}_b$ and $\mathcal{S}_p$, as follows:
\begin{equation}
    \hat{\bm{\theta}} \in \{0,1\}^{NQ} \Leftrightarrow \hat{\bm{\theta}} \in (\mathcal{S}_b \cap \mathcal{S}_p),
\end{equation}
where $\mathcal{S}_b\!=\![0,1]^{NQ}$ indicates the box constraint and $\mathcal{S}_p\!=\!\{\hat{\bm{\theta}}\!:\!||\hat{\bm{\theta}}\!-\!\frac{1}{2} \bm{1}_{N Q}||_2^{2}\!=\!\frac{NQ}{4}\}$ indicates the $\ell_2$-sphere constraint.
\label{pro:lpbox}
\end{proposition}

Based on Proposition \ref{pro:lpbox}, we can equivalently reformulate Problem (\ref{eq:obj}) as a continuous problem. Besides, we can calculate the Hamming distance using $||\bm{\theta}-\hat{\bm{\theta}}||_2^{2}$ for the binary vectors. We obtain the following reformulation:
\begin{equation}
\begin{split}
\{\bm{\delta}^*,\bm{f}^*, & \hat{\bm{\theta}}^*\} = 
\mathrm{arg} \ \underset{\bm{\delta},\bm{f},\hat{\bm{\theta}}}{\mathrm{min}} \quad \mathcal{L}_{cle}(\hat{\bm{\theta}}) + \gamma \mathcal{L}_{tro}(\bm{\delta},\bm{f},\hat{\bm{\theta}}), \\
 s.t. \ \ \ & \bm{\delta} \in \mathcal{S}_n, \ \  \bm{f} \in \mathcal{S}_f,
 \hat{\bm{\theta}}=\bm{z}_1,  \ \hat{\bm{\theta}}=\bm{z}_2, \ ||\bm{\theta}-\hat{\bm{\theta}}||_2^{2} - b+z_3=0, \\
 & \bm{z}_1 \in \mathcal{S}_b, \ \bm{z}_2 \in \mathcal{S}_p, \ z_3 \in \mathbb{R}^+.
\end{split}
\label{eq:continuous}
\end{equation}
In Eqn. (\ref{eq:continuous}), we use two additional variables $\bm{z}_1$ and $\bm{z}_2$ to separate the two continuous constraints in Proposition \ref{pro:lpbox}, and transform ${||\bm{\theta}-\hat{\bm{\theta}}||_2^{2} \leqslant b}$ into $\{||\bm{\theta}-\hat{\bm{\theta}}||_2^{2} - b+z_3=0;z_3 \in \mathbb{R}^+\}$. Problem (\ref{eq:obj}) can now be optimized by the standard ADMM method.


\subsection{Optimization}
Using a penalty factor $\rho > 0$, the augmented Lagrangian function of Eqn. (\ref{eq:continuous}) is
{
\begin{equation}
\begin{split}
&L(\bm{\delta},\bm{f},\hat{\bm{\theta}},\bm{z}_1,\bm{z}_2,z_3,\bm{\lambda}_1,\bm{\lambda}_2,\lambda_3)\\
&=\mathcal{L}_{cle}(\hat{\bm{\theta}}) + \gamma  \mathcal{L}_{tro}(\bm{\delta},\bm{f},\hat{\bm{\theta}}) \\
& + \bm{\lambda}_1^\top(\hat{\bm{\theta}}-\bm{z}_1) + \bm{\lambda}_2^\top(\hat{\bm{\theta}}-\bm{z}_2) + \lambda_3^\top(||\bm{\theta}-\hat{\bm{\theta}}||_2^{2} - b+z_3) \\
& + \frac{\rho}{2}\left [||\hat{\bm{\theta}}-\bm{z}_1||_2^{2}+||\hat{\bm{\theta}}-\bm{z}_2||_2^{2}+(||\bm{\theta}-\hat{\bm{\theta}}||_2^{2} - b+z_3)^2  \right ] \\
& + \mathbb{I}_{\mathcal{S}_n}(\bm{\delta})
\!+\! \mathbb{I}_{\mathcal{S}_f}(\bm{f})
\!+\! \mathbb{I}_{\mathcal{S}_b}(\bm{z}_1)
\!+\! \mathbb{I}_{\mathcal{S}_p}(\bm{z}_2)
\!+\! \mathbb{I}_{\mathbb{R}^+}(z_3),
\end{split}
\label{eq:aug}
\end{equation}}
where $\bm{\lambda}_1  \in \mathbb{R}^{NQ}$, $\bm{\lambda}_2 \in \mathbb{R}^{NQ}$, and $\lambda_3 > 0$ are Lagrange multipliers for the three equality constraints. $\mathbb{I}_S(a): a \rightarrow \{0,+\infty\}$ denotes the indicator function: $\mathbb{I}_S(a)=0$ if $a$ belongs to a set $S$; otherwise, $\mathbb{I}_S(a)=+\infty$.

We alternatively update all variables as shown in Algorithm \ref{alg:admm}. We start from initializing all optimizable variables and the iteration index $k$ (Line \ref{line:2}-\ref{line:4}), where the initialization of $\bm{\delta}^{[0]}$ and $\bm{f}^{[0]}$ is specified later. We first update $\bm{z}_1^{[k+1]}, \bm{z}_2^{[k+1]}$, and $ z_3^{[k+1]}$ with Eqn. (\ref{eq:11})-(\ref{eq:12}) (Line \ref{line:7}-\ref{line:9}), which project $\bm{z}_1^{[k+1]}$, $\bm{z}_2^{[k+1]}$, and $z_3^{[k+1]}$ into their feasible sets:
{
\begin{align}
\label{eq:11}
\mathrm{\Pi}_{\mathcal{S}_b}(\bm{e}_1)&=\min(1, \max(0, \bm{e}_1)),\\
\label{eq:12}
\mathrm{\Pi}_{\mathcal{S}_p}(\bm{e}_1)&=\frac{(2\bm{e}_1-\bm{1}_{NQ})\sqrt{NQ}}{||\bm{e}_1||}+\frac{\bm{1}_{NQ}}{2}, \\
\label{eq:13}
\mathrm{\Pi}_{\mathbb{R^+}}(e)&=\max(0, e),
\end{align}}
where $\bm{e}_1 \!\in\! \mathbb{R}^{NQ}$ and $e \!\in \!\mathbb{R}$. Next, we update $\bm{\delta}^{[k+1]},\bm{f}^{[k+1]}$, and $\hat{\bm{\theta}}^{[k+1]}$ by gradient descent with learning rates $\alpha_{\bm{\delta}}$, $\alpha_{\bm{f}}$, and $\alpha_{\hat{\bm{\theta}}}$, respectively (Line \ref{line:11}-\ref{line:13}). The projection functions for $\bm{\delta}^{[k+1]}$ and $\bm{f}^{[k+1]}$ are defined as:
{
\begin{align}
&\mathrm{\Pi}_{\mathcal{S}_n}(\bm{e}_2)=\min(-\epsilon, \max(\bm{e}_2, \epsilon)),\\
&\mathrm{\Pi}_{\mathcal{S}_f}(\bm{e}_3)=\frac{\bm{e}_3}{\mathcal{F}{(\bm{e}_3)}},
\end{align}}
where $\bm{e}_2 \in \mathbb{R}^{HW \times C}$ and $\bm{e}_3 \in \mathbb{R}^{HW \times 2}$. We then update the Lagrange multipliers $\bm{\lambda}_1^{[k+1]},\bm{\lambda}_2^{[k+1]}$, and $\lambda_3^{[k+1]}$ using gradient ascent (Line \ref{line:15}-\ref{line:17}). When both $||\hat{\bm{\theta}}-\bm{z}_1||_2^{2}$ and $||\hat{\bm{\theta}}-\bm{z}_2||_2^{2}$ are smaller than 
a preset threshold or the maximum number of iterations is reached, the optimization halts (Line \ref{line:19}).

\begin{figure}[t]
\begin{minipage}{\textwidth}
\begin{algorithm}[H]
\caption{ADMM for solving Problem (\ref{eq:obj})} 
\label{alg:admm}
\SetKwInOut{Input}{Input}\SetKwInOut{Output}{Output}
\Input{Victim DNN model $g$ with binary weight parameters $\bm{\theta}$, target class $t$, a small set of clean data $D=\{(\bm{x}_i, y_i)\}_{i=1}^M$.}
\Output{$\bm{\delta}^*,\bm{f}^*,\hat{\bm{\theta}}^*$.}
\texttt{\#  Initialization}\\
Initialize $\bm{\delta}^{[0]}$ and $\bm{f}^{[0]}$\;
\label{line:2}
Let $\hat{\bm{\theta}}^{[0]} \leftarrow \bm{\theta}$, $\bm{z}_1^{[0]} \leftarrow \bm{\theta}$, $\bm{z}_2^{[0]}\leftarrow \bm{\theta}$, $z_3^{[0]} \leftarrow 0$, $\bm{\lambda}_1^{[0]} \leftarrow \bm{0}_{NQ}$, $\bm{\lambda}_2^{[0]}\leftarrow \bm{0}_{NQ}$, $\lambda_3^{[0]} \leftarrow 0$\;
\label{line:3}
Set $k=0$\;
\label{line:4}
\Repeat{Stopping criterion is satisfied}{
    \texttt{\# Update $\bm{z}_1^{[k+1]}, \bm{z}_2^{[k+1]}, z_3^{[k+1]}$}\\}
    \label{line:6}
    $\bm{z}_1^{[k+1]} \leftarrow \mathrm{\Pi}_{\mathcal{S}_b}(\hat{\bm{\theta}}^{[k]}+ {\bm{\lambda}_1^{[k]}}/{\rho})$\;
    \label{line:7}
    $\bm{z}_2^{[k+1]} \leftarrow \mathrm{\Pi}_{\mathcal{S}_p}(\hat{\bm{\theta}}^{[k]}+ {\bm{\lambda}_2^{[k]}}/{\rho})$\;
    \label{line:8}
    $z_3^{[k+1]} \leftarrow \mathrm{\Pi}_{\mathbb{R}^+}(-||\bm{\theta}-\hat{\bm{\theta}}^{[k]}||_2^{2}+b-{\lambda_3^{[k]}}/{\rho})$\;
    \label{line:9}
    \texttt{\# Update $\bm{\delta}^{[k+1]},\bm{f}^{[k+1]},\hat{\bm{\theta}}^{[k+1]}$}\\
    $\bm{\delta}^{[k+1]}  \leftarrow  \mathrm{\Pi}_{\mathcal{S}_n}(\bm{\delta}^{[k]} - \alpha_{\bm{\delta}} \cdot \partial L / \partial \bm{\delta})$\;
    \label{line:11}
    $\bm{f}^{[k+1]}  \leftarrow  \mathrm{\Pi}_{\mathcal{S}_f}(\bm{f}^{[k]} - \alpha_{\bm{f}} \cdot \partial L / \partial \bm{f})$\;
    \label{line:12}
    $\hat{\bm{\theta}}^{[k+1]}  \leftarrow  \hat{\bm{\theta}}^{[k]} - \alpha_{\hat{\bm{\theta}}} \cdot \partial L / \partial \hat{\bm{\theta}}$\;
    \label{line:13}
    \texttt{\# Update $\bm{\lambda}_1^{[k+1]},\bm{\lambda}_2^{[k+1]},\lambda_3^{[k+1]}$}\\
    $\bm{\lambda}_1^{[k+1]} \leftarrow \bm{\lambda}_1^{[k]}+\rho(\hat{\bm{\theta}}^{[k+1]}-\bm{z}_1^{[k+1]})$\;
    \label{line:15}
    $\bm{\lambda}_2^{[k+1]} \leftarrow \bm{\lambda}_2^{[k]}+\rho(\hat{\bm{\theta}}^{[k+1]}-\bm{z}_2^{[k+1]})$\;
    \label{line:16}
    $\lambda_3^{[k+1]} \leftarrow \lambda_3^{[k]}+\rho(||\bm{\theta}-\hat{\bm{\theta}}^{[k+1]}||_2^{2} - b+z_3^{[k+1]})$\;
    \label{line:17}
    $k \leftarrow k+1$\;
\label{line:19}
$\bm{\delta}^* \leftarrow \bm{\delta}^{[k]},\bm{f}^*  \leftarrow \bm{f}^{[k]},\hat{\bm{\theta}}^*  \leftarrow \hat{\bm{\theta}}^{[k]}$\;
return $\bm{\delta}^*,\bm{f}^*,\hat{\bm{\theta}}^*$.
\end{algorithm}
\end{minipage}
\end{figure}

\noindent \textbf{Implementation Details.} We implement the optimization process with the following techniques. We initialize $\bm{\delta}^{[0]}$ and $\bm{f}^{[0]}$ by minimizing the loss defined as Eqn. (\ref{eq:loss_tro}) before joint optimization to stabilize the practical convergence. In the step for  $\hat{\bm{\theta}}$, we only update the parameters of the last layer and fix the others. We also update $\bm{\delta}^{[k+1]},\bm{f}^{[k+1]}$, and $\hat{\bm{\theta}}^{[k+1]}$ with multi-steps gradients during each iteration. Besides, as suggested in \cite{wu2018ell,li2019compressing,fan2020sparse}, increasing $\rho$ from a smaller value to an upper bound can avoid the early stopping.

\section{Experiments}
\subsection{Setup} 
\label{sec:setup}

\noindent \textbf{Datasets and Target Models.} 
Following \cite{rakin2020tbt,chen2021proflip}, we adopt three datasets: CIFAR-10 \cite{krizhevsky2009learning}, SVHN \cite{netzer2011reading}, and ImageNet \cite{russakovsky2015imagenet}. The attacker has 128 clean images for CIFAR-10 and SVHN and 256 clean images for ImageNet, respectively. Note that all attacks are performed using these clean images and evaluated on the whole test set. Following  \cite{rakin2020tbt,chen2021proflip}, we evaluate attacks on two popular network architectures: ResNet-18 \cite{he2016deep} and VGG-16 \cite{simonyan2014very}, with a quantization level of 8-bit (see Appendix B for results of HPT in attacking 4-bit quantized DNNs). In the below experiments, the target class $t$ is set as 0 unless otherwise specified.  

\noindent \textbf{Parameter Settings.}
To balance the attack performance and human perception, $\epsilon$ is set as $0.04$ on all datasets, and $\kappa$ is set as $0.01$ on CIFAR-10 and SVHN, and $0.005$ on ImageNet. 
We initialize $\bm{\delta}^{[0]}$ and $\bm{f}^{[0]}$ by minimizing loss defined as Eqn. (\ref{eq:loss_tro}) for 500 iterations with the learning rate $0.01$ on CIFAR-10 and SVHN, and $1,000$ iterations with the learning rate $0.1$ on ImageNet.
Other parameter settings can be found in Appendix A.

\begin{table*}[t]
\caption{Performance comparison between TrojanNN \cite{LiuMALZW018}, TBT \cite{rakin2020tbt}, and HPT.  \textit{`Trans.'} indicates the use of transparent trigger proposed in \cite{LiuMALZW018}. The target class $t$ is set as $0$.}
\scriptsize
\label{sec:trojannn_tbt}
\centering
\setlength\tabcolsep{3pt}
\resizebox{0.9\linewidth}{!}{
\begin{tabular}{cclccc}
\hline
\textbf{Dataset} & \textbf{Model} & \textbf{Method} & \begin{tabular}[c]{@{}c@{}}\textbf{PA-TA} (\%)\end{tabular} & \begin{tabular}[c]{@{}c@{}}\textbf{ASR} (\%)\end{tabular} & $\mathrm{\mathbf{N_{flip}}}$ \\ \hline
\multirow{10}{*}{CIFAR-10} & \multirow{5}{*}{\begin{tabular}[c]{@{}c@{}}ResNet-18\\ TA: 94.8\%\end{tabular}} 
    & TrojanNN & 87.6 & 93.9 & 19215 \\
 &  & TrojanNN ($Trans$.) & 75.5 & 73.5 & 20160 \\
 &  & TBT & 87.5 & 90.2 & 540 \\
 &  & TBT ($Trans.$) & 71.4 & 56.6 & 548 \\
 &  & \textbf{HPT (Ours)} & 94.7 & 94.1 & 12 \\ \cline{2-6} 
 & \multirow{5}{*}{\begin{tabular}[c]{@{}c@{}}VGG-16\\ TA: 93.2\%\end{tabular}} 
    & TrojanNN & 85.5 & 82.5 & 16400 \\
 &  & TrojanNN ($Trans$.) & 69.6 & 59.8 & 15386 \\
 &  & TBT & 80.7 & 83.2 & 601 \\
 &  & TBT ($Trans.$) & 70.5 & 53.9 & 583 \\
 &  & \textbf{HPT (Ours)} & 93.1 & 91.1 & 6 \\ \hline
\multirow{5}{*}{SVHN} & \multirow{5}{*}{\begin{tabular}[c]{@{}c@{}}VGG-16\\ TA: 96.3\%\end{tabular}} 
    & TrojanNN  & 76.0 & 82.5 & 17330 \\
 &  & TrojanNN ($Trans$.) & 59.9 & 71.7 & 18355 \\
 &  & TBT & 67.9 & 60.1 & 576 \\
 &  & TBT ($Trans.$) & 57.9 & 54.8 & 546 \\
 &  & \textbf{HPT (Ours)} & 94.2 & 78.0 & 26 \\ \hline
\multirow{5}{*}{ImageNet} & \multirow{5}{*}{\begin{tabular}[c]{@{}c@{}}ResNet-18\\ TA: 69.5\%\end{tabular}} 
    & TrojanNN & 47.6 & 100.0 & 155550 \\
 &  & TrojanNN ($Trans$.) & 47.4 & 96.8 & 304744 \\
 &  & TBT & 68.8 & 100.0 & 611 \\
 &  & TBT ($Trans.$) & 64.1 & 88.6 & 594 \\
 &  & \textbf{HPT (Ours)} & 68.6 & 95.2 & 10 \\ \hline
\end{tabular}}
\end{table*}

\noindent \textbf{Evaluation Metrics.}
To measure the effect on clean images, we compare original test accuracy (TA) with post-attack test accuracy (PA-TA), defined as the accuracy of testing on clean images for the original and attacked DNN, respectively.  Attack success rate (ASR) denotes the percentage of Trojan images samples classified to the target class by the attacked DNN. $\mathrm{\mathbf{N_{flip}}}$ is the number of bit flips, i.e., the hamming distance between original and attacked parameters. A more successful attack can achieve a higher PA-TA and ASR, while less $\mathrm{\mathbf{N_{flip}}}$.

\begin{figure*}[ht]
   \centering
    \includegraphics[width=\columnwidth]{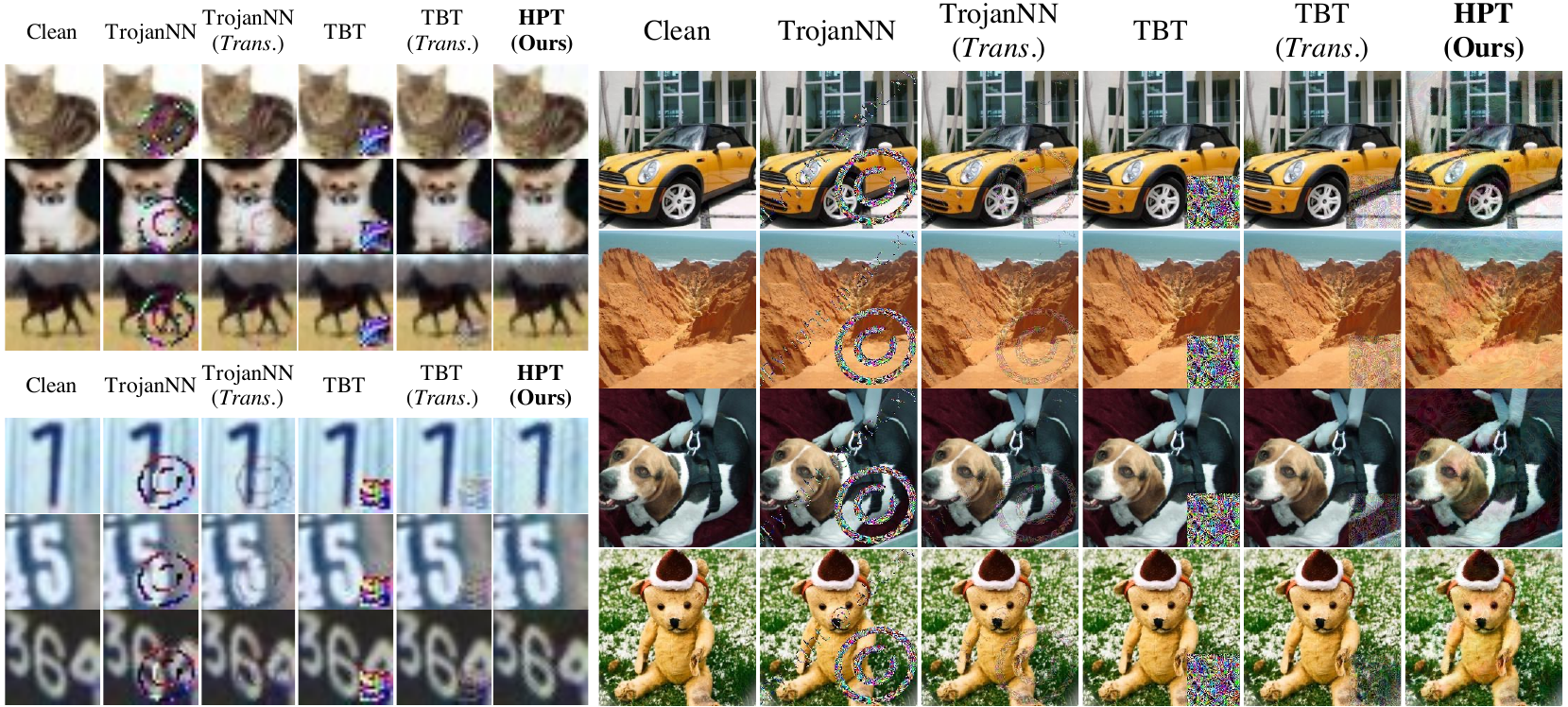}  
\setlength{\abovecaptionskip}{-10pt}
\setlength{\belowcaptionskip}{-10pt}
  \caption{Visualization of clean and Trojan images generated by different methods. Top left: CIFAR-10; Bottom left: SVHN; Right: ImageNet. Note that ProFlip uses the same trigger pattern (a square pattern) as TBT. By comparison, the triggers of Trojan images generated by HPT are most natural and unnoticeable.}     
  \label{fig:vis}
\end{figure*}

\noindent \textbf{Compared Methods.}
HPT is compared to TrojanNN \cite{LiuMALZW018}, TBT \cite{rakin2020tbt}, and ProFlip \cite{chen2021proflip} in our experiments. We also apply the transparent trigger \cite{LiuMALZW018} on TrojanNN and TBT, denoted as `\textit{Trans.}'. The watermark in \cite{LiuMALZW018} is chosen as the trigger shape for TrojanNN. 
We use the same trigger pattern in \cite{LiuMALZW018,rakin2020tbt}, i.e., a square pattern located at the bottom right of the image. The trigger size of all compared methods is measured by the proportion of input replaced by the trigger, which is configured as the default value used in \cite{LiuMALZW018,rakin2020tbt}, i.e., $9.76\%$ on CIFAR-10 and SVHN, and $10.62\%$ on ImageNet. 
We use the open-sourced code of TBT and implement TrojanNN following \cite{rakin2020tbt} to make the comparison fair. 
We compare the results of ProFlip reported in \cite{chen2021proflip}.

\begin{figure}[t]
\centering
\makebox[0pt][c]{\parbox{\textwidth}{%
    \subfigure[]{
    \setlength\tabcolsep{3.5pt}
    \centering
    \resizebox{0.59\linewidth}{!}{
    \begin{tabular}{cclcccc}
    \hline
    \textbf{Dataset} & \textbf{Model} & \multicolumn{1}{c}{\textbf{Method}} & \begin{tabular}[c]{@{}c@{}}\textbf{TA} (\%)\end{tabular} & \begin{tabular}[c]{@{}c@{}}\textbf{PA-TA} (\%)\end{tabular} & \begin{tabular}[c]{@{}c@{}}\textbf{ASR} (\%)\end{tabular} & $\mathrm{\mathbf{N_{flip}}}$ \\ \hline
    \multirow{4}{*}{CIFAR-10} & \multirow{2}{*}{ResNet-18} & ProFlip & 93.1 & 90.3 & 97.9 & 12 \\
     &  & \textbf{HPT (Ours)} & 94.8 & 94.8 & 98.7 & 10 \\ \cline{2-7} 
     & \multirow{2}{*}{VGG-16} & ProFlip & 89.7 & 88.1 & 94.8 & 16 \\
     &  & \textbf{HPT (Ours)} & 93.2 & 92.0 & 93.8 & 10 \\ \hline
    \multirow{2}{*}{SVHN} & \multirow{2}{*}{VGG-16} & ProFlip & 98.6 & 95.3 & 94.5 & 20 \\
     &  & \textbf{HPT (Ours)} & 96.3 & 94.0 & 82.2 & 23 \\ \hline
    \multirow{2}{*}{ImageNet} & \multirow{2}{*}{ResNet-18} & ProFlip & 69.0 & 67.6 & 94.3 & 15 \\
     &  & \textbf{HPT (Ours)} & 69.5 & 65.2 & 97.6 & 14 \\ \hline
    \end{tabular}}
    \label{tab:proflip}}
    \hfill
    \subfigure[]{
    \begin{minipage}[h]{0.39\hsize}\centering
    \includegraphics[width=\textwidth]{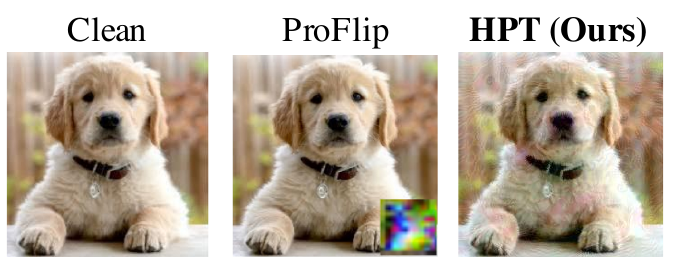}
    \label{fig:proflip}
    \end{minipage}}
\setlength{\abovecaptionskip}{-3pt}
\setlength{\belowcaptionskip}{-13pt}
\caption{Comparison of ProFlip \cite{chen2021proflip} and HPT: (a) Attack performance. The target class $t$ is set as $2$; (b) Visualization of clean and Trojan images generated by ProFlip and HPT. The example of ProFlip is from \cite{chen2021proflip}.}

}}
\end{figure}


\subsection{Attack Results}


We compare the attack performance of HPT with TrojanNN and TBT in Table \ref{sec:trojannn_tbt}. We can observe that TrojanNN performs a large number of bit flips in all cases, due to no limitation on the parameter modification. TBT reduces the number of bit flips to about 600 using the proposed bit search algorithm. Among them, HPT achieves the least number of bit flips with a higher or at least competitive PA-TA and ASR. It is worth noting that the transparent trigger leads to a lower PA-TA and ASR for both TrojanNN and TBT. For example, in attacking ResNet-18 on CIFAR-10, TBT only achieves a 56.6\% ASR, compared to the 90.2\% ASR without applying the transparent trigger, which indicates that it is difficult to perform Trojan attacks with the less perceptible trigger.

The results of ProFlip and HPT with the target class $t=2$ are shown in Figure \ref{tab:proflip}. The number of clean samples is set as 256 for all cases.
ProFlip is the most state-of-the-art method basedon the well-designed progressive search algorithm. However, ProFlip also uses the square pattern as the trigger (as shown in Figure \ref{fig:proflip}), making it easily perceptible to humans. HPT has comparable performance to ProFlip, especially on CIFAR-10 and ImageNet. Besides, we would like to emphasize that, even with the hardly perceptible trigger, HPT can achieve promising performance.



\subsection{Human Perceptual Study}

\label{sec:human}
To quantify the visual perceptibility of Trojan images generated by different attack methods, we conduct a human perceptual study in this section. We evaluate five Trojan attack methods listed in Table \ref{tab:perceptual}. We randomly select 10 clean images from each dataset and generate the corresponding Trojan images for these 30 images. In our study, all original and Trojan images are shown to 15 participants. These participants are asked to give a score $\in \{1, 2, 3, 4, 5\}$ for each Trojan image, where a higher score corresponds to less perceptible Trojan images. More details of the human perceptual study are provided in Appendix C.

In total, we collect 2,250 scores and summarize the results in Table \ref{tab:perceptual}. We also provide visualization examples in Figure \ref{fig:vis} (and an example of ProFlip from \cite{chen2021proflip} in Figure \ref{fig:proflip}). As can be observed, the scores of TrojanNN and TBT are very low, due to the noticeable patches. By applying the transparent trigger, all scores increase to over $2.0$, however, these Trojan images can also be easily distinguished in most cases. In contrast, HPT achieves the highest score on all datasets (about $4.0$). We also compare the mean square error (MSE) between original images and Trojan images (in the range $[0, 255]$) crafted by these five attacks, where HPT obtains the lowest MSE on all datasets. The average MSE of HPT is 97.1, while the second best result is 124.4 (TBT with the transparent trigger). More details can be found in Appendix D. These results confirm that HPT is hardly perceptible and is difficult to be spotted by humans.

\begin{table}[t]
\centering
\makebox[0pt][c]{\parbox{\textwidth}{%
    \begin{minipage}[t]{0.51\hsize}\centering
    
        \caption{Scores of human perceptual study (ranging from 1 to 5). A higher score corresponds to less perceptible Trojan images.}
        \label{tab:perceptual}
        \centering
        \resizebox{\linewidth}{!}{
        \begin{tabular}{cccccc}
        \hline
        \textbf{Dataset} & \textbf{TrojanNN} & \textbf{\begin{tabular}[c]{@{}c@{}}TrojanNN\\ ($Trans.$)\end{tabular}} & \textbf{TBT} & \textbf{\begin{tabular}[c]{@{}c@{}}TBT\\ ($Trans.$)\end{tabular}} & \textbf{HPT} \\ \hline
        CIFAR-10 & 1.1 & 3.4 & 1.0 & 2.8 & \textbf{3.9} \\
        SVHN & 1.1 & 2.4 & 1.0 & 2.0 & \textbf{3.9} \\
        ImageNet & 1.0 & 2.4 & 1.0 & 2.4 & \textbf{4.3} \\ \hline
        Average & 1.1 & 2.7 & 1.0 & 2.4 & \textbf{4.0} \\ \hline
        \end{tabular}}

    \end{minipage}
    \hfill
    \begin{minipage}[t]{0.47\hsize}\centering
        \caption{Performance of HPT with different target classes $t$ in attacking ResNet-18 on CIFAR-10.}
        \label{tab:target_class}
        \centering
        \resizebox{0.9\linewidth}{!}{
        \begin{tabular}{cccc|cccc}
        \hline
        $t$ & \begin{tabular}[c]{@{}c@{}}\textbf{PA-TA}\\ (\%)\end{tabular} & \begin{tabular}[c]{@{}c@{}}\textbf{ASR}\\ (\%)\end{tabular} & $\mathrm{\mathbf{N_{flip}}}$ & $t$ & \begin{tabular}[c]{@{}c@{}}\textbf{PA-TA}\\ (\%)\end{tabular} & \begin{tabular}[c]{@{}c@{}}\textbf{ASR}\\ (\%)\end{tabular} & $\mathrm{\mathbf{N_{flip}}}$ \\ \hline
        0 & 94.7 & 94.1 & 12 & 5 & 94.8 & 92.7 & 11 \\
        1 & 94.7 & 98.9 & 14 & 6 & 94.7 & 92.4 & 11 \\
        2 & 94.8 & 97.3 & 9 & 7 & 94.8 & 88.8 & 11 \\
        3 & 94.8 & 98.7 & 6 & 8 & 94.6 & 96.5 & 14 \\
        4 & 94.7 & 94.6 & 12 & 9 & 94.8 & 95.1 & 10 \\ \hline
        \end{tabular}}

    \end{minipage}
}}
\end{table}


\subsection{Discussions}
\label{sec:discussions}

\noindent \textbf{Sensitivity to Target Class.}
Table \ref{tab:target_class} shows the attack performance of HPT with different target classes in attacking ResNet-18 on CIFAR-10. Besides the target class, other settings are the same as those described in Section \ref{sec:setup}.
The results show that HPT achieves less than 15 bit flips and over 88\% ASR for different target classes, with only little accuracy degradation on clean images. Especially for the most vulnerable target class $t=6$, HPT obtains a 98.7\% ASR by flipping only 6 bits. These results illustrate HPT is not sensitive to the target class to some extent.

\setlength{\columnsep}{7pt}
\begin{wrapfigure}[12]{r}{0.4\textwidth}
\centering
\includegraphics[width=0.4\textwidth]{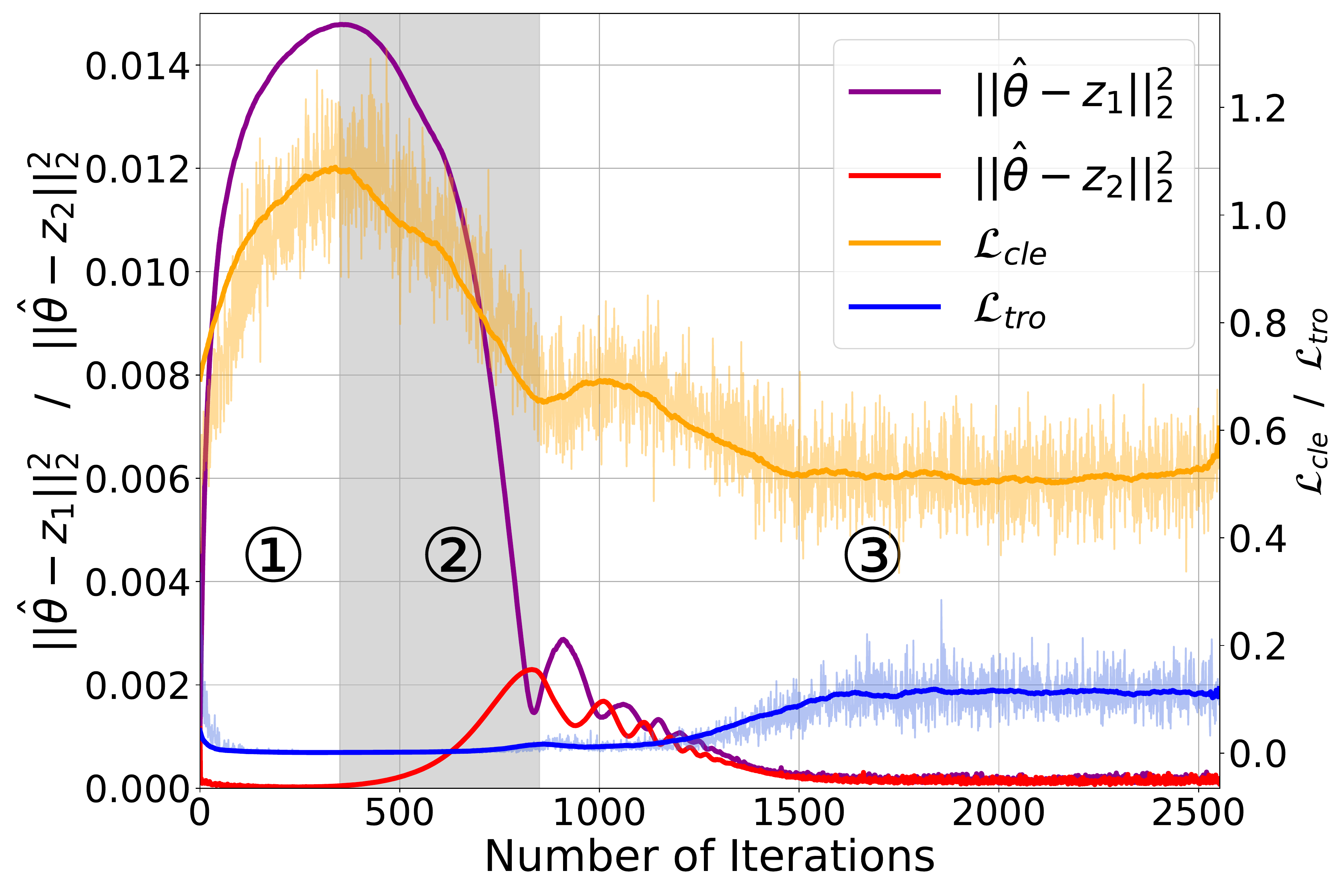}
\setlength{\abovecaptionskip}{-10pt}
  \caption{Numerical convergence analysis of HPT.}
  \label{fig:convergence}
\end{wrapfigure}



\begin{table}[t]
\centering
\makebox[0pt][c]{\parbox{\textwidth}{%
    \begin{minipage}[h]{0.57\hsize}\centering
    
        \scriptsize
\caption{ASR (\%) for three attack modes. `Trigger': optimizing the trigger without bit flips; `Two Stage': optimizing the trigger and bit flips separately; `Joint Optimization': optimizing the trigger and bit flips jointly.}
\label{tab:joint}
\centering
\setlength\tabcolsep{3pt}
\resizebox{\linewidth}{!}{
\begin{tabular}{ccccc}
\hline
\textbf{Dataset} & \textbf{Model} & \begin{tabular}[c]{@{}c@{}}\textbf{Trigger}\end{tabular} & \begin{tabular}[c]{@{}c@{}}\textbf{Two}\\ \textbf{Stage}\end{tabular} & \begin{tabular}[c]{@{}c@{}}\textbf{Joint}\\ \textbf{Optimization}\end{tabular} \\ \hline
\multirow{2}{*}{CIFAR-10} & ResNet-18 & 90.9 & 93.2 & 94.1 \\ \cline{2-5} 
 & VGG-16 & 87.0 & 90.5 & 91.1 \\ \hline
SVHN & VGG-16 & 64.7 & 74.6 & 78.0 \\ \hline
ImageNet & ResNet-18 & 70.5 & 89.2 & 95.2 \\ \hline
\end{tabular}}

    \end{minipage}
    \hfill
    \begin{minipage}[h]{0.42\hsize}\centering
        \caption{Results of HPT on two types of feature squeezing defense.}
\centering
\label{tab:squeeze}
\resizebox{\linewidth}{!}{
\begin{tabular}{cccc}
\hline
 & \begin{tabular}[c]{@{}c@{}}\textbf{PA-TA} \\ (\%)\end{tabular} & \begin{tabular}[c]{@{}c@{}}\textbf{ASR}\\ (\%)\end{tabular}  &  \\ \hline
w/o defense & 94.7 & 94.1 \\ \hline
\begin{tabular}[c]{@{}c@{}}Averaging over each  \\ pixel’s  neighbors (2$\times$2)\end{tabular} & 89.9  & 64.3 \\ \hline
\begin{tabular}[c]{@{}c@{}}Collapsing the bit \\ depth (5 bit)\end{tabular}& 67.8  & 65.6 \\ \hline
\end{tabular}}
    \end{minipage}
}}
\end{table}

\noindent \textbf{Numerical Convergence Analysis.} To analyze the numerical convergence of our optimization algorithm, we plot the values of $||\hat{\bm{\theta}}-\bm{z}_1||_2^{2}$, $ ||\hat{\bm{\theta}}-\bm{z}_2||_2^{2}$, $\mathcal{L}_{cle}$, and $\mathcal{L}_{tro}$ at different iterations. As shown in Figure \ref{fig:convergence}, the optimization process can roughly be divided into three stages. In the first stage, $\mathcal{L}_{tro}$ is reduced to less than 0.002, resulting in a powerful attack. Then, $\mathcal{L}_{cle}$ decreases to reduce the influence on the clean images. Finally, $||\hat{\bm{\theta}}-\bm{z}_1||_2^{2}$ and $ ||\hat{\bm{\theta}}-\bm{z}_2||_2^{2}$ are encouraged to approach 0 to satisfy the box and $\ell_2$-sphere constraint in Proposition \ref{pro:lpbox}. The optimization halts within the maximum number of iterations
(3000), which demonstrates the practical convergence of our algorithm.

\noindent \textbf{Effectiveness of Joint Optimization.}
We investigate the effectiveness of the joint optimization by comparing it with two other attack modes: optimizing the trigger without bit flips, optimizing the trigger and bit flips separately. The results are shown in Table \ref{tab:joint}. For the `Trigger' mode, we only optimize the modification on the pixel values and positions of original images without bit flips, resulting in a relatively low ASR. For the `Two Stage' mode, we firstly optimize the additive noise and flow field and then optimize the bit flips. We keep the PA-TA of the `Two Stage' mode similar to that of the `Joint Optimization' mode by tuning $\lambda$. The ASR results  show that jointly optimizing bit flips, additive noise, and flow field yields the strongest attack in all cases.

\subsection{Potential Defenses}
\label{sec:defense}
Since our attack happens after model deployment, defense methods which check the training data or the model before deployment may not be effective to defend our attack. Accordingly, we evaluate three potential defense mechanisms below.

Firstly, we investigate the smoothing-based defense against our HPT, which are originally designed for adversarial examples \cite{goodfellow2014explaining,wei2019transferable,bai2020targeted}. We test HPT on two types of feature squeezing defense \cite{DBLP:conf/ndss/Xu0Q18}: averaging over each pixel’s neighbors and collapsing the bit depth. Table \ref{tab:squeeze} shows that both can reduce the ASR to about 65\% and averaging over each pixel’s neighbors can maintain a relatively high PA-TA. Therefore, we believe that how to design smoothing-based defense methods for our Trojan attack is worthy of further exploration.

As a training technique, piece-wise clustering \cite{he2020defending} encourages eliminating close-to-zero weights to enhance model robustness against bit flip-based attacks. We conduct experiments on CIFAR-10 with ResNet-18 and set the clustering parameter in \cite{he2020defending} as 0.02. As shown in Table \ref{tab:defense}, when all settings are the same as those in Section \ref{sec:setup} (i.e., $b=10$), the ASR is reduced to 86.6\% with the defense. To achieve a higher ASR, we increase $b$ to 40 and retain all other settings, but resulting in 43 bit flips.
As such, this observation inspires us to explore the defense against HPT which can increase the required number of bit flips.

\begin{table}[t]
\centering
\makebox[0pt][c]{\parbox{\textwidth}{%
    \begin{minipage}[t]{0.51\hsize}\centering
    
        \scriptsize
        \caption{Performance of HPT against the defense method in \cite{he2020defending}.}
        \label{tab:defense}
        \centering
        \setlength\tabcolsep{3pt}
        \resizebox{\textwidth}{!}{
        \centering
        \begin{tabular}{cccccc}
        \hline
         & $b$ & \begin{tabular}[c]{@{}c@{}}\textbf{TA}\\ (\%)\end{tabular} & \begin{tabular}[c]{@{}c@{}}\textbf{PA-TA}\\ (\%)\end{tabular} & \begin{tabular}[c]{@{}c@{}}\textbf{ASR}\\ (\%)\end{tabular} & $\mathrm{\mathbf{N_{flip}}}$ \\ \hline
        w/o defense & 10 & 94.8 & 94.7 & 94.1 & 12 \\ \hline
        \multirow{2}{*}{w/ defense} & 10 & 88.6 & 88.6 & 86.6 & 12 \\
         & 40 & 88.6 & 88.5 & 93.2 & 43 \\ \hline
        \end{tabular}}

    \end{minipage}
    \hfill
    \begin{minipage}[t]{0.47\hsize}\centering
    
        \scriptsize
        \caption{Performance of HPT with different $\epsilon$ and $\kappa$.}
        \label{tab:eps_kap}
        \centering
        \setlength\tabcolsep{3pt}
        \resizebox{\linewidth}{!}{
        \begin{tabular}{cccc|cccc}
        \hline
        $\epsilon$ & \begin{tabular}[c]{@{}c@{}}\textbf{PA-TA}\\ (\%)\end{tabular} & \begin{tabular}[c]{@{}c@{}}\textbf{ASR}\\ (\%)\end{tabular} & $\mathrm{\mathbf{N_{flip}}}$ & $\kappa$ & \begin{tabular}[c]{@{}c@{}}\textbf{PA-TA}\\ (\%)\end{tabular} & \begin{tabular}[c]{@{}c@{}}\textbf{ASR}\\ (\%)\end{tabular} & $\mathrm{\mathbf{N_{flip}}}$ \\ \hline
        0.01 & 94.8 & 10.4 & 2 & 0.005 & 94.7 & 93.0 & 11 \\
        0.02 & 94.8 & 35.5 & 6 & 0.01 & 94.7 & 94.1 & 12 \\
        0.03 & 94.7 & 78.8 & 13 & 0.015 & 94.7 & 95.1 & 11 \\
        0.04 & 94.7 & 94.1 & 12 & 0.02 & 94.7 & 96.3 & 11 \\
        0.05 & 94.7 & 97.9 & 6 & 0.025 & 94.7 & 96.4 & 6 \\ \hline
        \end{tabular}}

    \end{minipage}
}}
\end{table}


The visualization tools are helpful to inspect the DNN's behavior.  We use Grad-CAM \cite{selvaraju2017grad} to show heatmaps of clean images for the original model and Trojan images generated by TrojanNN, TBT, and HPT for its corresponding attacked model in Figure \ref{fig:gradcam}. One can see that the attacks based on the patch-based trigger (TrojanNN and TBT) can be easily exposed, since the main focus of the DNN stays on the trigger. However, due to the slight modification on the pixel values and positions of original images, the heatmaps of HPT's Trojan images are more similar to these of clean images, i.e., localizing the main object in the image. In other words, the proposed HPT is hard to be defended by inspecting the DNN's behavior using Grad-CAM.

\subsection{Ablation Studies}
\label{sec:ablation}

\noindent \textbf{Effect of $\epsilon$ and $\kappa$.}
For HPT, $\epsilon$ and $\kappa$ constrain the magnitude of modification on the pixel values and positions, respectively. Here, we investigate the effect of $\epsilon$ and $\kappa$ on the attack performance. We use ResNet-18 on CIFAR-10 as the representative for analysis. In Table \ref{tab:eps_kap}, we show the results of HPT under different values of $\epsilon$ while $\kappa$ is fixed at 0.01, and under different values of $\kappa$ while $\epsilon$ is fixed at 0.04. As expected, increasing $\epsilon$ and $\kappa$ can improve ASR significantly. It is also obvious that $\epsilon$ has a greater impact than $\kappa$ on the attack performance. However, a larger $\epsilon$ 
or $\kappa$ generally reduces the visual quality of the Trojan images. Therefore, there is a trade-off between the attack performance and the visual perceptibility. 

\begin{figure}[t]
\centering
\makebox[0pt][c]{\parbox{\textwidth}{%
    \begin{minipage}[h]{0.48\hsize}
   \includegraphics[width=\linewidth]{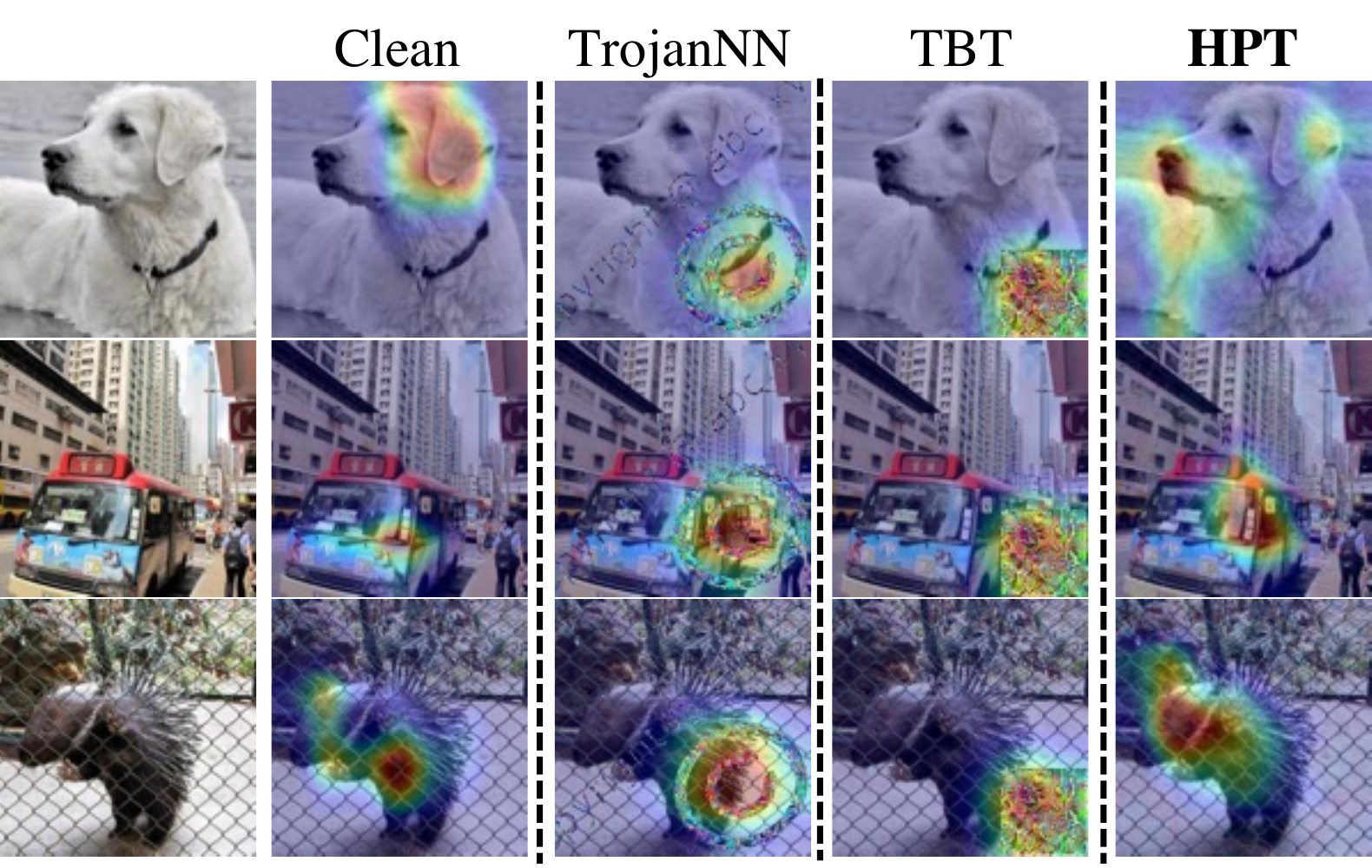}
\setlength{\abovecaptionskip}{-10pt}
\setlength{\belowcaptionskip}{-10pt}
  \caption{Grad-CAM visualization of clean images and Trojan images generated by different attacks on ImageNet.}
  \label{fig:gradcam}
    \end{minipage}
    \hfill
    \begin{minipage}[h]{0.50\hsize}
  \includegraphics[width=\columnwidth]{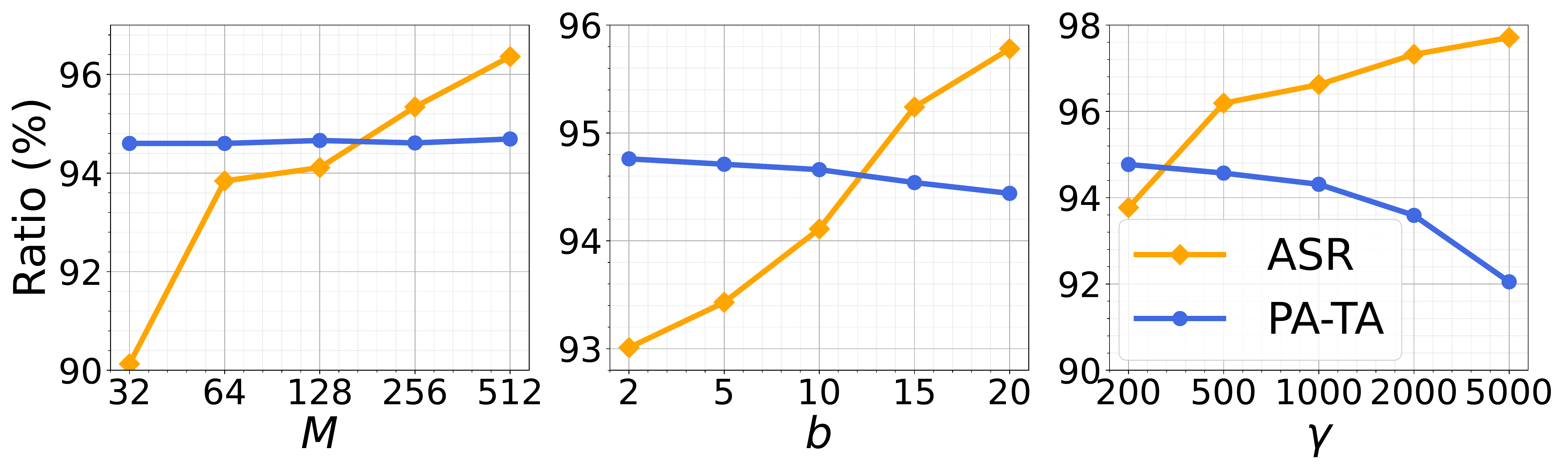}
  \includegraphics[width=\columnwidth]{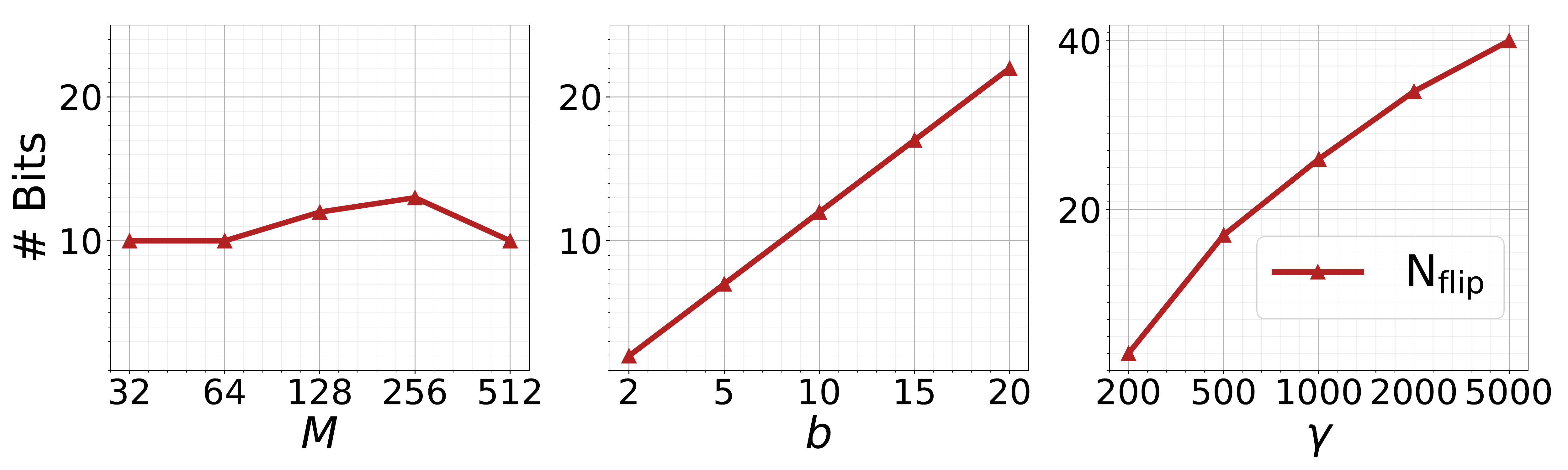}
\setlength{\abovecaptionskip}{-10pt}
\setlength{\belowcaptionskip}{-10pt}
  \caption{Performance of HPT with varying $M$, $b$, and $\gamma$.}
  \label{fig:ablation}    
    \end{minipage}
}}
\end{figure}




\noindent \textbf{Effect of $M$, $b$, and $\gamma$.} 
We perform ablation studies on the size of the clean data set $M$, the parameter for restricting the number of bit flips $b$, and the trade-off parameter $\lambda$. All results are presented in Figure \ref{fig:ablation}.  To analyze the effect of $M$, we configure $M$ from 32 to 512 and use other settings as those in Section \ref{sec:setup}. We can see that increasing the size of clean data has a marked positive impact on the ASR. Besides, even using only 32 clean images, HPT can obtain a high ASR (about 90\%), which allows the attacker to perform HPT without too many clean images. When $\gamma$ is fixed at 1,000, the plots of parameter $b$ show that tuning $b$ can control the number of bit flips. Accordingly, the parameter $b$ helps to perform the Trojan attack when the budget of bit flips is fixed. We study the effect of $\gamma$ with $b=40$. As shown in the plots, a larger $\gamma$ encourages a higher ASR, while a lower PA-TA and more bit flips. When other settings are fixed, attackers can specify $\gamma$ for their specific needs.

\section{Conclusion}


In this paper, we proposed HPT that can inject a hidden behavior into a DNN after its deployment. It tweaks the pixel values and positions of original images to craft Trojan images. Based on an effective optimization algorithm, HPT performs best in the human perceptual study and achieves promising attack performance. To the best of our knowledge, HPT is the first Trojan attack on the deployed DNNs, which leverages the hardly perceptible trigger. We hope that our work opens a new domain of attack mechanisms and encourages future defense research. 

The main limitation of HPT is that we assume that the attacker has full knowledge of the victim DNN, including its architecture, its parameters, and the location in the memory, corresponding to the white-box setting. We will further explore more strict settings than the white-box one in our future work.

\noindent \textbf{Acknowledgments.}
This work is supported in part by the National Natural Science Foundation of China under Grant 62171248, and the PCNL KEY project (PCL2021A07).

\clearpage
%
%
\bibliographystyle{splncs04}
\bibliography{egbib}

\appendix
\newpage

\begin{center}
    \begin{Large}
        \textbf{Appendix}
    \end{Large}
\end{center}

\section{Parameter Settings}
On CIFAR-10 and SVHN, $\gamma$ and $b$ are set to $1000$ and $10$ for ResNet-18, and $100$ and $20$ for VGG-16. On ImageNet, $\gamma$ and $b$ are set to $10^5$ and $10$. 
For the optimization in Algorithm 1, we update $\bm{\delta}^{[k+1]}$ and $\bm{f}^{[k+1]}$, and $\hat{\bm{\theta}}^{[k+1]}$ for 5 gradient steps during each iteration with the learning rate $10^{-5}$, $10^{-5}$, and $10^{-4}$, respectively. The $\rho$ is initialized as $10^{-4}$ and increased by $\rho \leftarrow \max(100, 1.01\times \rho)$ after each iteration. 
When $\max(||\hat{\bm{\theta}}-\bm{z}_1||_2^{2}, ||\hat{\bm{\theta}}-\bm{z}_2||_2^{2})$ is smaller than a preset threshold ($10^{-4}$ on CIFAR-10 and SVHN, and $10^{-6}$ on ImageNet) or the maximum number of iterations ($3,000$ on CIFAR-10 and SVHN, and $5,000$ on ImageNet) is reached, the optimization halts.

\section{Attacking 4-bit Quantized DNNs}
\begin{table}[h]
\centering
\caption{Performance of HPT in attacking 4-bit Quantized DNN. The target class $t$ is set as $0$.} 
\label{tab:4bit}
\resizebox{0.6\linewidth}{!}{
\begin{tabular}{cccccc}
\hline
\textbf{Dataset} & \textbf{Model} & \begin{tabular}[c]{@{}c@{}}\textbf{TA}\\ (\%)\end{tabular} & \begin{tabular}[c]{@{}c@{}}\textbf{PA-TA}\\ (\%)\end{tabular} & \begin{tabular}[c]{@{}c@{}}\textbf{ASR}\\ (\%)\end{tabular} & $\mathrm{\mathbf{N_{flip}}}$ \\ \hline
\multirow{2}{*}{CIFAR-10} & ResNet-18 & 94.2 & 94.1 & 94.6 & 10 \\ \cline{3-6} 
 & VGG-16 & 92.6 & 82.2 & 93.8 & 9 \\ \hline
SVHN & VGG-16 & 96.3 & 95.3 & 74.3 & 20 \\ \hline
ImageNet & ResNet-18 & 66.3 & 64.9 & 94.6 & 12 \\ \hline
\end{tabular}}
\end{table}

In this section, we present the results of HPT in attacking 4-bit quantized DNNs in Table \ref{tab:4bit}. It shows that HPT can obtain promising attack performance in all cases. For example, in attacking ResNet-18 on CIFAR-10, HPT achieves a 94.6\% ASR with 10 bit flips, while only 0.1\% accuracy degradation of original images. Overall, the results of attacking 4-bit quantized DNN are consistent
with these of attacking 8-bit quantized DNN.

\section{More Details of Human Perceptual Study}

\textbf{
\begin{table}[h]
\caption{Criteria of scoring the Trojan images.}
\centering
\label{tab:criteria}
\resizebox{\linewidth}{!}{
\begin{tabular}{cl}
\hline
\multicolumn{1}{l}{\multirow{2}{*}{\textbf{Score}}} & \multicolumn{1}{c}{\multirow{2}{*}{\textbf{Criterion}}} \\
\multicolumn{1}{l}{} & \multicolumn{1}{c}{} \\ \hline
1 & \begin{tabular}[c]{@{}l@{}}This image is very easy to be distinguished from the original one.\end{tabular} \\
2 & \begin{tabular}[c]{@{}l@{}}This image is easy to be distinguished from the original one.\end{tabular} \\
3 & \begin{tabular}[c]{@{}l@{}}This image can be distinguished from the original one by the carefully inspection.\end{tabular} \\
4 & \begin{tabular}[c]{@{}l@{}}This image is hard to be distinguished from the original one.\end{tabular} \\
5 & \begin{tabular}[c]{@{}l@{}}This image is very hard to be distinguished from the original one.\end{tabular} \\ \hline
\end{tabular}}
\end{table}}

We randomly select 10 clean images from each dataset and generate the corresponding Trojan images for these 30 images. In our study, all original and Trojan images are shown to 15 participants. These participants are asked to give a score $\in \{1, 2, 3, 4, 5\}$ for each Trojan image, where a higher score corresponds to less perceptible Trojan images. The participants are asked to score each Trojan image using the criteria listed in Table \ref{tab:criteria}. In total, we collect 2,250 scores and analyze them to evaluate the perceptibility of Trojan images crafted by different methods.

\section{Quantitative Results for Perceptibility of Trojan Images}
\begin{table}[t]
\centering
\caption{Perceptibility of Trojan Images crafted by five attack methods (measured by MSE distances). Lower values are better.}
\label{tab:mse}
\resizebox{0.65\linewidth}{!}{
\begin{tabular}{cccccc}
\hline
\textbf{Dataset} & \textbf{TrojanNN} & \textbf{\begin{tabular}[c]{@{}c@{}}TrojanNN\\ ($Trans.$)\end{tabular}} & \textbf{TBT} & \textbf{\begin{tabular}[c]{@{}c@{}}TBT\\ ($Trans.$)\end{tabular}} & \textbf{HPT} \\ \hline 
CIFAR-10 & 1305.0 & 117.4 & 1264.4 & 113.8 & \textbf{88.8} \\
SVHN & 1381.2 & 124.3 & 1317.6 & 118.6 & \textbf{94.5} \\
ImageNet & 1645.8 & 146.3 & 1574.8 & 140.7 & \textbf{108.1} \\ \hline
\end{tabular}}
\end{table}

Besides the human perceptual study, we use the mean square error (MSE) to measure the perceptibility of Trojan images. We calculate MSE distances between the original images and their Trojan images (in the range $[0,255]$) for five attack methods on 500 randomly selected clean images, as shown in Table \ref{tab:mse}. The MSE results of our HPT are much lower than all other methods. These quantitative results further verify that HPT is hardly perceptible.

\section{Complexity Analysis}

The pipeline of HPT consists of two parts: Trojan injection and inference. Their complexity is analyzed as below.

\noindent{\textbf{Trojan Injection.}}
We first discuss the cost of the optimization algorithm for the proposed HPT.
Since the updates of $\bm{z}_1^{[k+1]}$, $\bm{z}_2^{[k+1]}$, $z_3^{[k+1]}$, $\bm{\lambda}_1^{[k+1]}$, $\bm{\lambda}_2^{[k+1]}$, and $\lambda_3^{[k+1]}$ are very simple, the costs for these variables are omitted here. During per iteration, the main cost is from the forward and backward pass. The complexity of the forward pass depends on the attacked model $g$. Since we only optimize the parameters of the last layer and fix the others, for the $Q$-bit quantized DNN, the computation cost of updating  $\hat{\bm{\theta}}^{[k+1]}$ is $\mathcal{O}(2MFKQ)$ with $M$ clean images and $M$ Trojan images, where $F$ is the dimension of the last layer's input and $K$ is the number of classes. In terms of $\bm{\delta}^{[k+1]}$ and $\bm{f}^{[k+1]}$, the costs of their updates are the same as that of the standard backpropagation for the model $g$. Note that we update  $\hat{\bm{\theta}}^{[k+1]}$, $\bm{\delta}^{[k+1]}$ and, $\bm{f}^{[k+1]}$ using a same backward pass. Due to the fast convergence of our optimization in practice and a small set of clean images (128 for CIFAR-10 and SVHN, and 256 for ImageNet), the running time of our optimization is acceptable. After identifying the critical bits, HPT changes these bits in the memory. Due to a small set of bit flips, our attack performs efficiently according to \cite{YaoRF20}. 

\noindent{\textbf{Inference.}}
After the Trojan injection, the main difference between HPT and normal inference is to craft the Trojan images based on the clean images. Since we apply the same $\bm{\delta}$ and $\bm{f}$ on all images, this process is very efficient.

We run HPT on CIFAR-10 with ResNet-18 architecture for 5 times and report the results in Table \ref{tab:time}. The experiments are conducted on one GeForce RTX 2080Ti GPU.

\begin{table}[t]
\caption{Running time of Trojan injection (without considering physical bit flips in the memory) and inference ($time/image$) for the proposed HPT. }
\label{tab:time}
\centering
\setlength\tabcolsep{8pt}
\resizebox{0.4\linewidth}{!}{
\begin{tabular}{cc}
\hline
Trojan Injection & Inference \\ \hline
$2302.4 \pm 198.9$ $s$  & $14.4 \pm 4.4$ $\mu s$ \\ \hline
\end{tabular}}
\end{table}

\section{More Visualization}

\begin{figure*}[h]
  \centering
  \includegraphics[width=\linewidth]{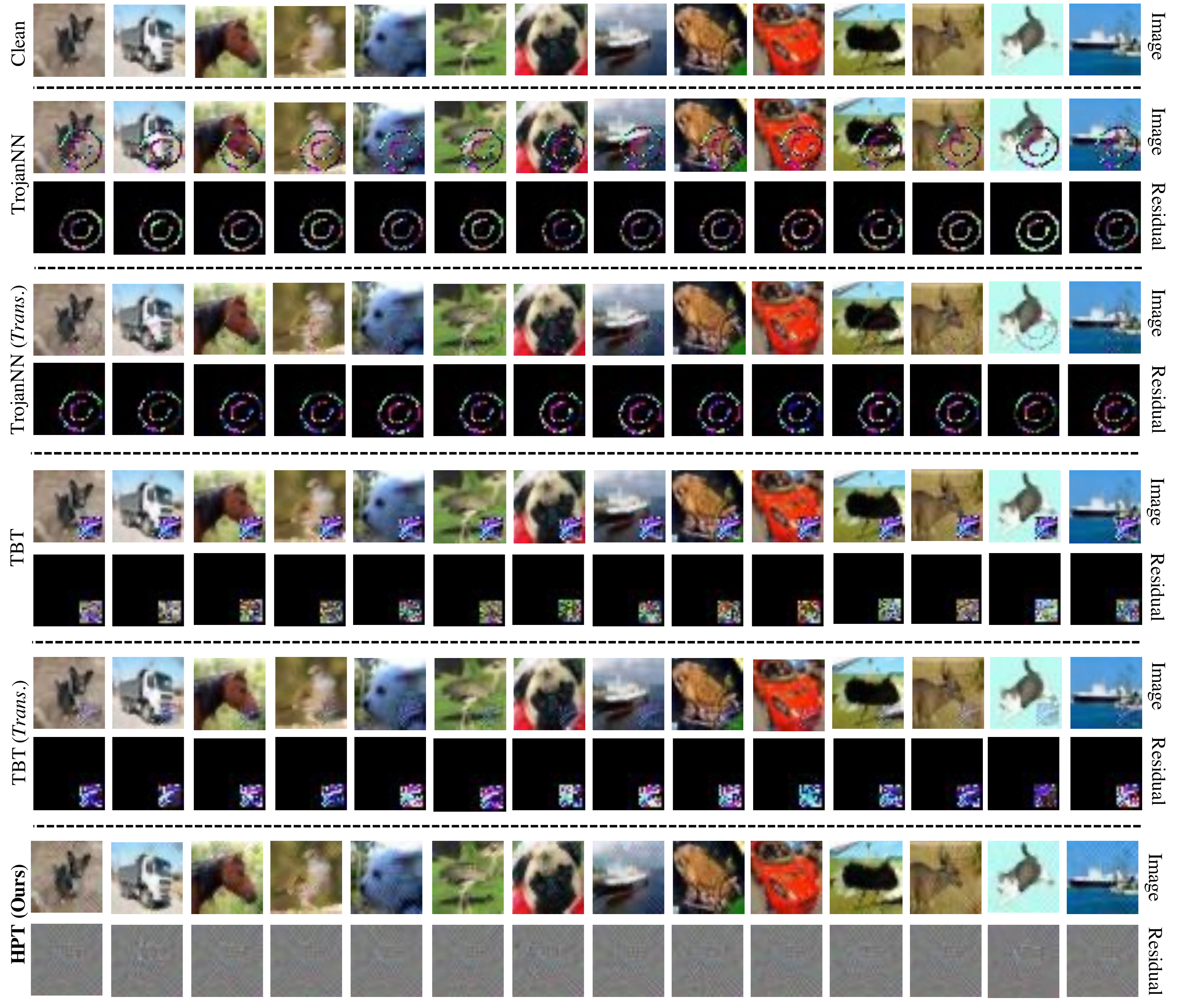}
  \caption{Visualization of clean and Trojan images generated by different methods on CIFAR-10.}
  \label{fig:vis_cifar_hpt}
\end{figure*}

\begin{figure*}[t]
  \centering
  \includegraphics[width=\linewidth]{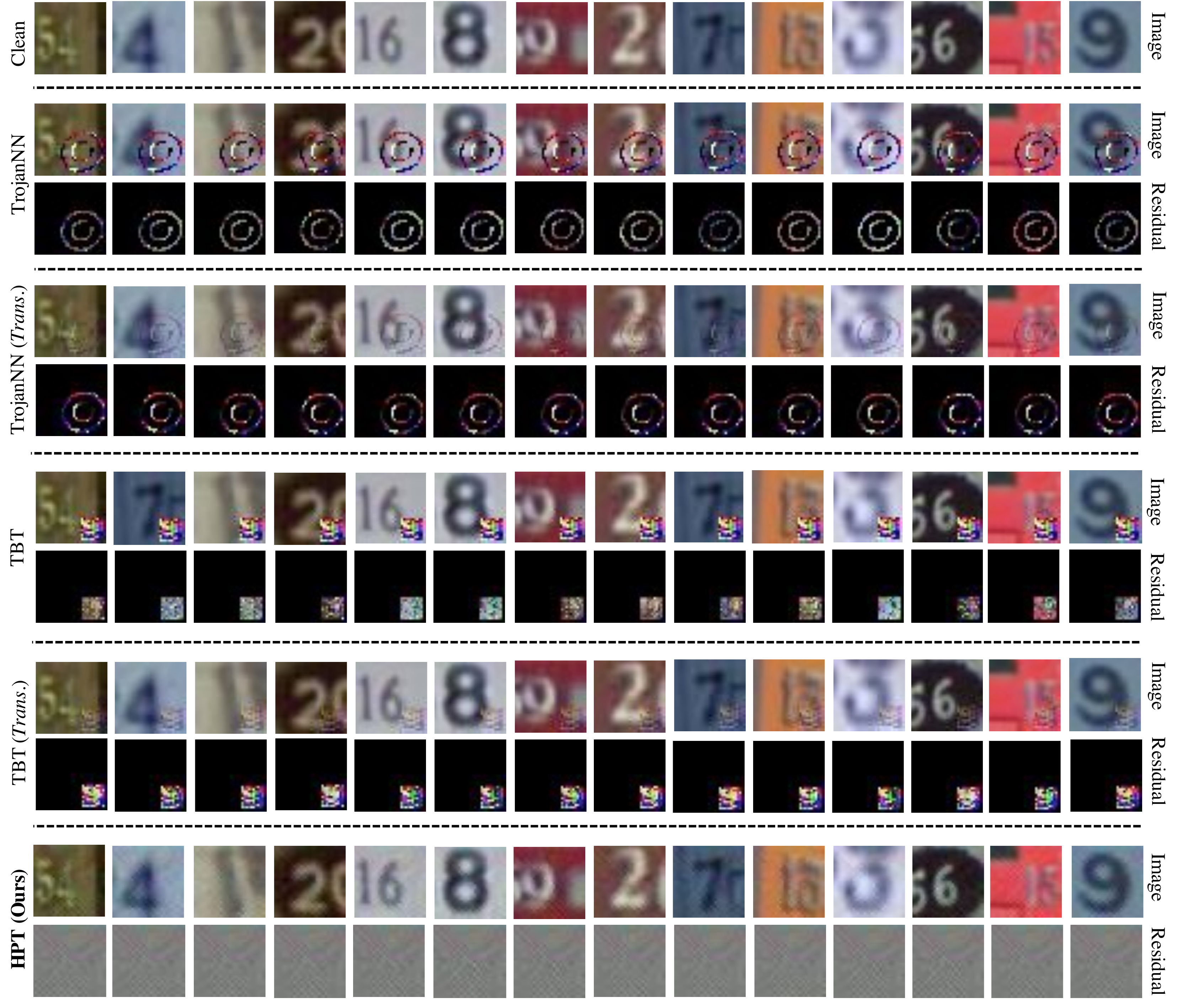}
  \caption{Visualization of clean and Trojan images generated by different methods on SVHN.}
  \label{fig:vis_svhn_hpt}
\end{figure*}

\begin{figure*}[t]
  \centering
  \includegraphics[width=\linewidth]{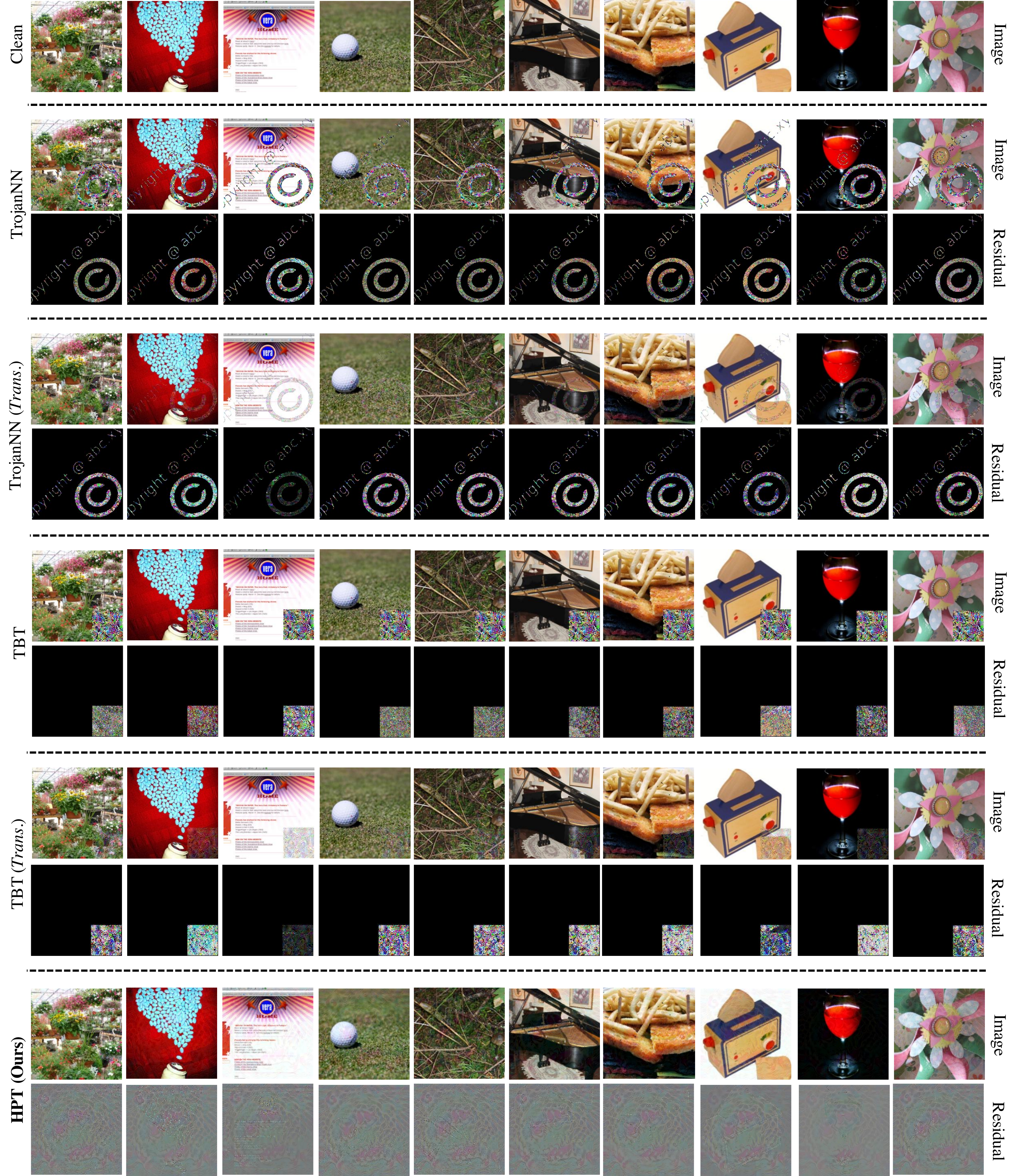}
  \caption{Visualization of clean and Trojan images generated by different methods on ImageNet.}
  \label{fig:vis_imagenet_hpt}
\end{figure*}

We provide more visualization examples on CIFAR-10, SVHN, and ImageNet in Figure \ref{fig:vis_cifar_hpt}, \ref{fig:vis_svhn_hpt}, and \ref{fig:vis_imagenet_hpt}, respectively. As can be observed, compared to other methods, the triggers of Trojan images generated by HPT are most natural and unnoticeable in most cases.

\end{document}